\documentclass[10pt,twocolumn,letterpaper]{article}

\usepackage{iccv}
\usepackage{times}
\usepackage{epsfig}
\usepackage{graphicx}
\usepackage{amsmath}
\usepackage{amssymb}

\usepackage{bm}
\usepackage{multirow}
\usepackage{booktabs}
\usepackage{enumerate}
\usepackage{subfig}
\usepackage{array}

\newcolumntype{x}[1]{>{\centering\arraybackslash}p{#1pt}}
\newcommand{\tablestyle}[2]{\setlength{\tabcolsep}{#1}\renewcommand{\arraystretch}{#2}\centering\footnotesize}
\usepackage[font=small]{caption}

\newcommand*\inlineimage[1]{\raisebox{-0.14\baselineskip}{\includegraphics[height=0.95\baselineskip]{#1}}}
\newcommand{\testpath}{\inlineimage{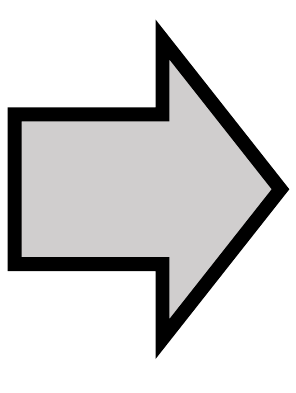}}
\newcommand{\trainpath}{\inlineimage{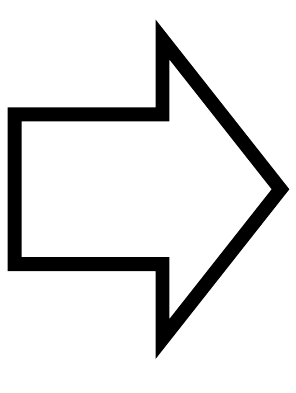}}

\usepackage{mathtools, nccmath}
\usepackage{footmisc}
\usepackage{amsthm}

\usepackage[pagebackref=true,breaklinks=true,letterpaper=true,colorlinks,bookmarks=false]{hyperref}

\iccvfinalcopy 

\def\iccvPaperID{1565} 

\ificcvfinal\fi
\begin{document}

\title{Counterfactual Critic Multi-Agent Training for Scene Graph Generation}

\author{Long Chen$^1$ \; Hanwang Zhang$^2$ \; Jun Xiao$^1$\thanks{Corresponding Author.} \; Xiangnan He$^3$ \; Shiliang Pu$^4$ \; Shih-Fu Chang$^5$ \\
$^1$DCD Lab, College of Computer Science and Technology, Zhejiang University \\ 
$^2$MReal Lab, Nanyang Technological University \; $^3$University of Science and Technology of China \\ 
$^4$Hikvision Research Institute \; $^5$DVMM Lab, Columbia University \\
}

\maketitle
\thispagestyle{empty}

\begin{abstract}
    Scene graphs --- objects as nodes and visual relationships as edges --- describe the whereabouts and interactions of objects in an image for comprehensive scene understanding. To generate coherent scene graphs, almost all existing methods exploit the fruitful visual context by modeling message passing among objects. For example, ``person'' on ``bike'' can help to determine the relationship ``ride'', which in turn contributes to the confidence of the two objects. However, we argue that the visual context is not properly learned by using the prevailing cross-entropy based supervised learning paradigm, which is not sensitive to graph inconsistency: errors at the hub or non-hub nodes should not be penalized equally. To this end, we propose a Counterfactual critic Multi-Agent Training (CMAT) approach. CMAT is a multi-agent policy gradient method that frames objects into cooperative agents, and then directly maximizes a graph-level metric as the reward. In particular, to assign the reward properly to each agent, CMAT uses a counterfactual baseline that disentangles the agent-specific reward by fixing the predictions of other agents. Extensive validations on the challenging Visual Genome benchmark show that CMAT achieves a state-of-the-art performance by significant gains under various settings and metrics.
\end{abstract}

\section{Introduction} \label{sec:1}

\begin{figure}[tbp]
	\centering
	\includegraphics[width=1\linewidth]{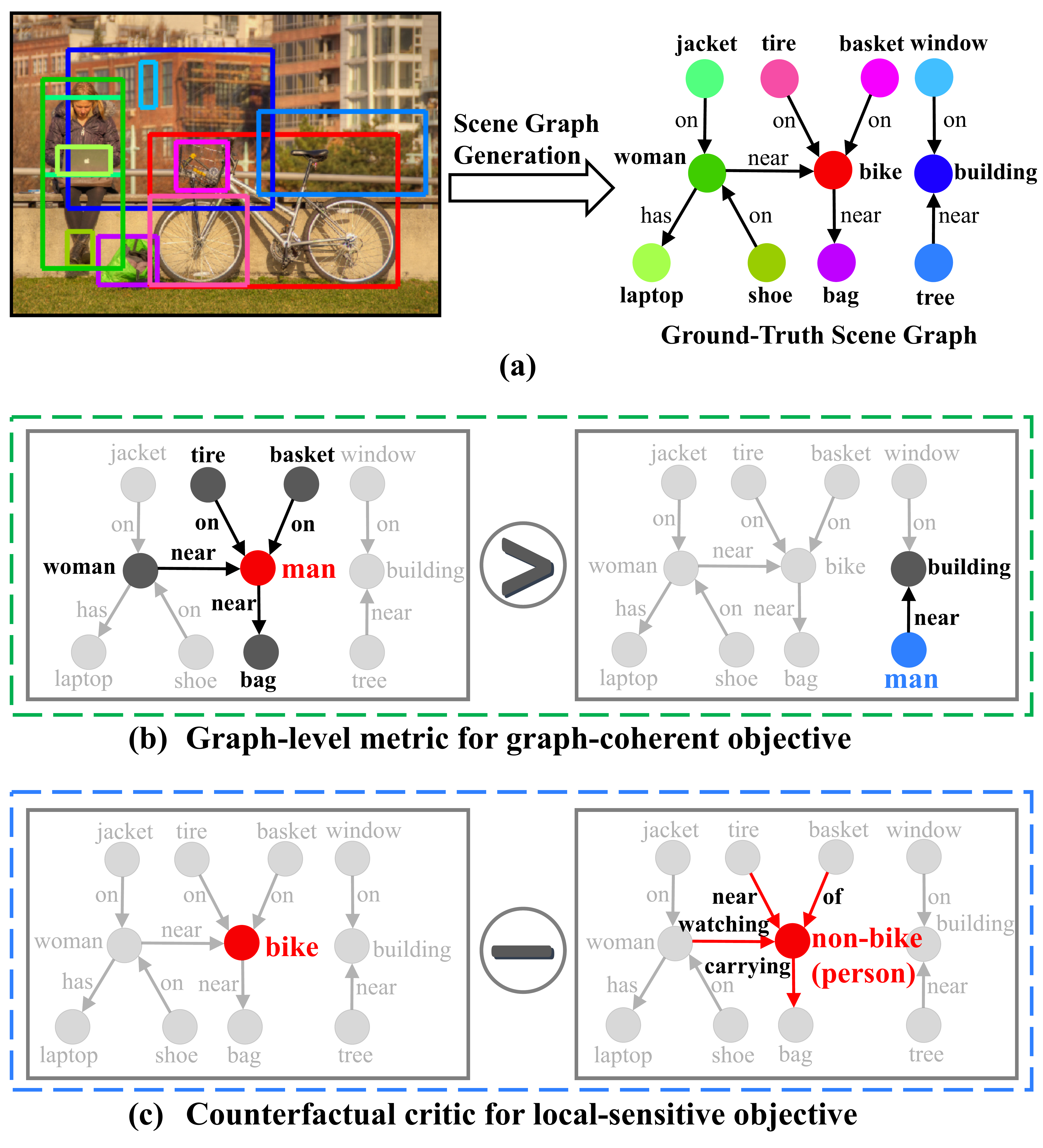}
	\vspace{-2.0em}
	\caption{(a) An input image and its ground-truth scene graph. (b) For graph-coherent objective, a graph-level metric will penalize the red node \textbf{more than} $(>)$ the blue one, even though both are misclassified as \texttt{man}. (c) For local-sensitive objective, the individual reward for predicting the red node as \texttt{bike} can be identified by \textbf{excluding} $(-)$ the reward  from \texttt{non-bike} predictions.} 
	\vspace{-1.0em}
\label{fig:1}
\end{figure}

Visual scene understanding, \eg, what and where the things and stuff are, and how they relate with each other, is one of the core tasks in computer vision. With the maturity of object detection~\cite{ren2015faster, liu2016ssd} and segmentation~\cite{long2015fully, he2017mask}, computers can recognize object categories, locations, and visual attributes well. However, scene understanding goes beyond the whereabouts of objects. A more crucial step is to infer their visual relationships --- together with the objects, they offer comprehensive and coherent visually-grounded knowledge, called \textbf{\emph{scene graphs}}~\cite{johnson2015image}. As shown in Figure~\ref{fig:1} (a), the nodes and edges in scene graphs are objects and visual relationships, respectively. Moreover, scene graph is an indispensable knowledge representation for many high-level vision tasks such as image captioning~\cite{yao2018exploring, yang2019auto, yang2019learning, Kim_2019_CVPR}, visual reasoning~\cite{shi2019explainable, Haurilet_2019_CVPR}, and VQA~\cite{norcliffe2018learning, hudson2019gqa}.

A straightforward solution for Scene Graph Generation (SGG) is in an independent fashion: detecting object bounding boxes by an existing object detector, and then predicting the object classes and their pairwise relationships separately~\cite{lu2016visual, zhang2017visual, yang2018shuffle, shang2017video}. However, these methods overlook the fruitful visual context, which offers a powerful inductive bias~\cite{divvala2009empirical} that helps object and relationship detection. For example in Figure~\ref{fig:1}, \texttt{window} and \texttt{building} usually co-occur within an image, and \texttt{near} is the most common relationship between \texttt{tree} and \texttt{building}; it is easy to infer that \texttt{?} is \texttt{building} from \texttt{window}-\texttt{on}-\texttt{?} or \texttt{tree}-\texttt{near}-\texttt{?}. Such intuition has been empirically shown benefits in boosting SGG~\cite{xu2017scene, dai2017detecting, li2017vip, li2017scene, li2018factorizable, yin2018zoom, jae2018tensorize, zellers2018neural, tang2019learning, gu2019scene, qi2019attentive, Wang_2019_CVPR, qian2019video}. More specifically, these methods use a conditional random field~\cite{zheng2015conditional} to model the joint distribution of nodes and edges, where the context is incorporated by message passing among the nodes through edges via a multi-step mean-field approximation~\cite{krahenbuhl2011efficient}; then, the model is optimized by the sum of cross-entropy (XE) losses of nodes (\eg, objects) and edges (\eg, relationships).

Nevertheless, the coherency of the visual context is not captured effectively by existing SGG methods due to the main reason: the XE based training objective is \textbf{not graph-coherent}. By ``graph-coherent'', we mean that the quality of the scene graph should be at the graph-level: the detected objects and relationships should be contextually consistent; however, the sum of XE losses of objects and relationships is essentially independent. To see the negative impact of this inconsistency, suppose that the red and the blue nodes are both misclassified in Figure~\ref{fig:1} (b). Based on the XE loss, the errors are penalized equally. However, the error of misclassifying the red node should be more severe than the blue one, as the red error will influence more nodes and edges than the blue one. Therefore, we need to use a graph-level metric such as Recall@K~\cite{lu2016visual} and SPICE~\cite{anderson2016spice}  to match the graph-coherent objective, which penalizes more for misclassifying important hub nodes than others. Meanwhile, the training objective of SGG should be \textbf{local-sensitive}. By ``local-sensitive'', we mean that the training objective is sensitive to the change of a single node. However, since the graph-coherent objective is a global pooling quantity, the individual contribution of the prediction of a node is lost. Thus, we need to design a disentangle mechanism to identify the individual contribution and provide an effective training signal for each local prediction.

In this paper, we propose a novel training paradigm: Counterfactual critic Multi-Agent Training (CMAT), to simultaneously meet the graph-coherent and local-sensitive requirements. Specifically, we design a novel communicative \textbf{\emph{multi-agent}} model, where the objects are viewed as cooperative agents to maximize the quality of the generated scene graph. The action of each agent is to predict its object class labels, and each agent can communicate with others using pairwise visual features. The communication retains the rich visual context in SGG. After several rounds of agent communication, a visual relationship model triggers the overall graph-level reward by comparing the generated scene graph with the ground-truth. 

\textbf{For the graph-coherent objective}, we directly define the objective as a graph-level reward (\eg, Recall@K or SPICE), and use policy gradient~\cite{sutton2000policy} to optimize the non-differentiable objective. In the view of Multi-Agent Reinforcement Learning (MARL)~\cite{tampuu2015multiagent, lowe2017multi}, especially the actor-critic methods~\cite{lowe2017multi}, the relationship model can be framed as a \textbf{\emph{critic}} and the object classification model serves as a policy network. \textbf{For the local-sensitive objective}, we subtract a counterfactual baseline~\cite{foerster2018counterfactual} from the graph-level reward by varying the target agent and fixing the others before feeding into the critic. As shown in Figure~\ref{fig:1} (c), to approximate the true influence of the red node acting as \texttt{bike}, we fix the predictions of the other nodes and replace the \texttt{bike} by \texttt{non-bike} (\eg, \texttt{person}, \texttt{boy}, and \texttt{car}), and see how such \textbf{\emph{counterfactual}} replacement affects the reward (\eg, the edges connecting their neighborhood are all wrong).

To better encode the visual context for more effective CMAT training, we design an efficient agent communication model, which discards the widely-used relationship nodes in existing message passing works~\cite{xu2017scene, li2017scene, jae2018tensorize, li2017vip, yin2018zoom, li2018factorizable}. Thanks to this design, we disentangle the agent communication (\ie, message passing) from the visual relationship detection, allowing the former to focus on modeling the visual context, and the latter, which is a communication consequence, to serve as the critic that guides the graph-coherent objective.

We demonstrate the effectiveness of CMAT on the challenging Visual Genome~\cite{krishna2017visual} benchmark. We observe consistent improvements across extensive ablations and achieve state-of-the-art performances on three standard tasks.

In summary, we make three contributions in this paper: 
\vspace{-0.6em}
\begin{enumerate}[1.]
\itemsep-0.4em
\item We propose a novel training paradigm: Counterfactual critic Multi-Agent Training (CMAT) for SGG. To the best of our knowledge, we are the first to formulate SGG as a cooperative multi-agent problem, which conforms to the graph-coherent nature of scene graphs.
\item We design a counterfactual critic that is effective for training because it makes the graph-level reward local-sensitive by identifying individual agent contributions. 
\item We design an efficient agent communication method that disentangles the relationship prediction from the visual context modeling, where the former is essentially a consequence of the latter.
\end{enumerate}

\begin{figure*}[htbp]
	\centering
	\includegraphics[width=1\linewidth]{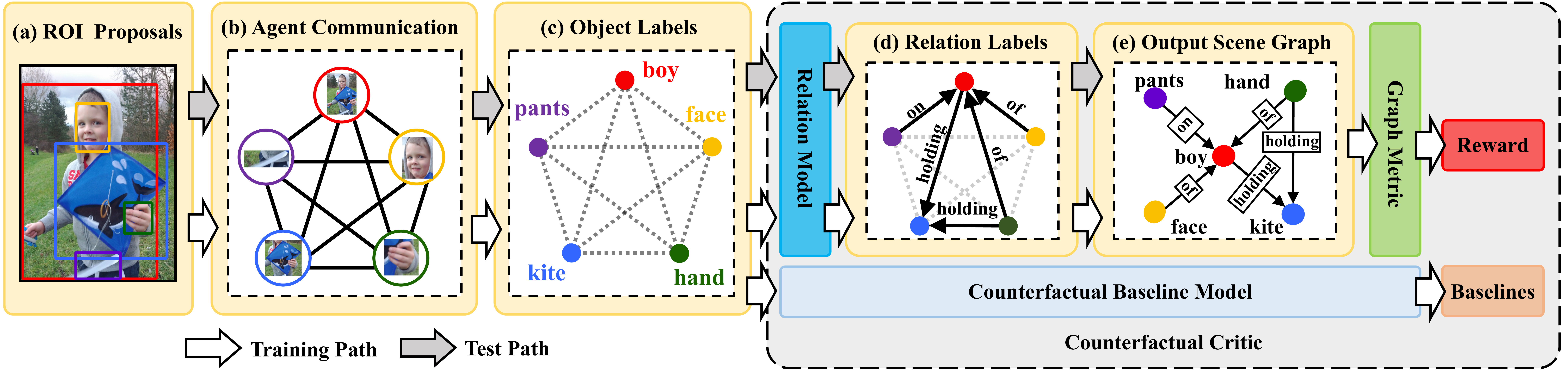}
	\vspace{-2.0em}
	\caption{The pipeline of CMAT framework. Given an image, the model uses RPN to propose object regions (a). Then, each object (agent) communicates with others to encode visual context (b). After agent communication, the model predicts class confidence for all objects. Based on the confidence, it selects (random or greedily sampling) object labels (c) and infers visual relationship of object pairs (d). Finally, it generates the scene graph (e). In the training stage, a counterfactual critic is used to calculate the individual contribution.}
    \vspace{-1em}
\label{fig:2}
\end{figure*}

\section{Related Work}

\noindent\textbf{Scene Graph Generation.}
Detecting visual relationships regains the community attention after the pioneering work by Lu~\etal~\cite{lu2016visual} and the advent of the first large-scale scene graph dataset by Krishna~\etal~\cite{krishna2017visual}. In the early stage, many SGG works focus on detecting objects and visual relations independently~\cite{lu2016visual, zhang2017visual, zhuang2017towards, zhu2018deep, zhang2017relationship}, but these independent inference models overlook the fruitful visual context. To benefit both object and relationship detection from visual context, recent SGG methods resort to the message passing mechanism~\cite{xu2017scene, dai2017detecting, li2017scene, li2018factorizable, yin2018zoom, jae2018tensorize, yang2018graph, tang2019learning, gu2019scene, qi2019attentive, Wang_2019_CVPR}. However, these methods fail to learn the visual context due to the conventional XE loss, which is not graph-level contextually consistent. Unlike previous methods, in this paper, we propose a CMAT model to simultaneously meet the graph-coherent and local-sensitive requirements.

\noindent\textbf{Multi-Agent Policy Gradient.}
Policy gradient is a type of method which can optimize non-differentiable objective. It had been well-studied in many scene understanding tasks like image captioning~\cite{ranzato2016sequence, ren2017deep, liu2017improved, rennie2017self, zhang2017actor, liu2018context}, VQA~\cite{hu2017learning, johnson2017inferring}, visual grounding~\cite{chen2017query, yu2017joint}, visual dialog~\cite{das2017learning}, and object detection~\cite{caicedo2015active, mathe2016reinforcement, jie2016tree}. Liang \etal~\cite{liang2017deep} used a DQN to formulate SGG as a single agent decision-making process. Different from these single agent policy gradient settings, we formulate SGG as a cooperative multi-agent decision problem, where the training objective is graph-level contextually consistent and conforms to the graph-coherent nature of scene graphs. Meanwhile, compared with many well-studied multi-agent game tasks~\cite{foerster2016learning, omidshafiei2017deep, foerster2017stabilising, tampuu2015multiagent}, the agent number (64 objects) and action sample space (151 object categories) in CMAT are much larger.

\section{Approach}
Given a set of predefined object classes $\mathcal{C}$ (including background) and visual relationship classes $\mathcal{R}$ (including non-relationship), we formally represent a scene graph $\mathcal{G} = \left\{ \mathcal{V}=\{(v_i, \bm{l}_i)\},     
\mathcal{E}=\{r_{ij}\} | i,j = 1...n \right\}$, where $\mathcal{V}$ and $\mathcal{E}$ denote the set of nodes and edges, respectively. $v_i \in \mathcal{C}$ is the object class of $i^{th}$ node, $\bm{l}_i \in \mathbb{R}^4$ is the location of $i^{th}$ node, and $r_{ij} \in \mathcal{R}$ is the visual relationship between $i^{th}$ and $j^{th}$ node. Scene Graph Generation (SGG) is to detect the coherent configuration for nodes and edges.

In this section, we first introduce the components of the CMAT (Section~\ref{sec:3.1}). Then, we demonstrate the details about the training objective of the CMAT (Section~\ref{sec:3.2}).

\subsection{SGG using Multi-Agent Communication} \label{sec:3.1}

We sequentially introduce the components of CMAT following the inference path (~\testpath~ path in Figure~\ref{fig:2}), including object proposals detection, agent communication, and visual relationship detection. 

\subsubsection{Object Proposals Detection} \label{sec:3.1.1}
\begin{small}
$$ \textbf{Input}: \text{IMAGE} \longmapsto \textbf{Output}: \{(\bm{l}_i, \bm{x}^0_i, \bm{s}^0_i)\} $$
\end{small}
We use Faster R-CNN~\cite{ren2015faster} as the object detector to extract a set of object proposals. Each proposal is associated with a location $\bm{l}_i$, a feature vector $\bm{x}^0_i$, and a class confidence $\bm{s}^0_i$. The superscript $0$ denotes the initial input for the following $T$-round agent communication. We follow previous works~\cite{xu2017scene, zellers2018neural} to fix all locations $\{\bm{l}_i\}$ as the final predictions. For simplicity, we will omit $\bm{l}_i$ in following sections.

\subsubsection{Agent Communication}
\begin{figure}[tbp]
	\centering
	\includegraphics[width=1\linewidth]{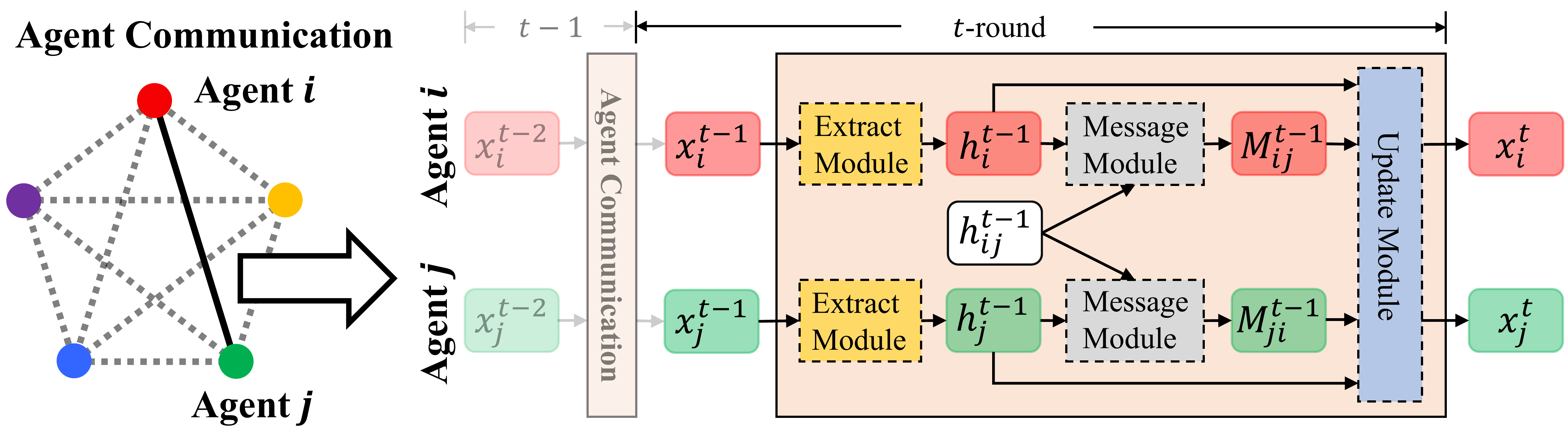}
	\caption{The illustration of agent communication (at time step t) between agent $i$ and agent $j$ (the \textcolor{red}{red} and \textcolor[rgb]{0.0, 0.7, 0.0}{green} node).}
\label{fig:3} \vspace{-0.5em}
\end{figure}

\begin{small}
$$\textbf{Input}: \{(\bm{x}^0_i, \bm{s}^0_i)\} \longmapsto \textbf{Output}: \{(\bm{x}^T_i, \bm{s}^T_i, \bm{h}^T_i)\}$$
\end{small}
Given the $n$ detected objects from the previous step, we regard each object as an agent and each agent will communicate with the others for $T$ rounds to encode the visual context. In each round of communication, as illustrated in Figure~\ref{fig:3}, there are three modules: \textbf{\texttt{extract}}, \textbf{\texttt{message}}, and \textbf{\texttt{update}}. These modules share parameters among all agents and time steps to reduce the model complexity. In the following, we introduce the details of these three modules.

\textbf{\texttt{Extract} Module.} 
The incarnation of the extract module is an LSTM, which encodes the agent interaction history and extracts the internal state of each agent. Specifically, for agent $i$ ($i^{th}$ object) at $t$-round ($0 < t \leqslant T$) communication:
\begin{equation}
\small
\begin{aligned}
    \bm{h}^{t}_i & = \text{LSTM}(\bm{h}^{t-1}_i, [\bm{x}^{t}_i, \bm{e}^{t-1}_i]), \\
    \bm{s}^t_i & = F_s(\bm{s}^{t-1}_i, \bm{h}^t_i), \; \bm{e}^t_i = F_e(\bm{s}^t_i), \\
\end{aligned}
\end{equation}
where $\bm{h}^t_i$ is the hidden state of LSTM (\ie, the internal state of agent). $\bm{x}^t_i$ is the time-step input feature and $\bm{s}^t_i$ is the object class confidence. The initialization of $\bm{x}^t_i$ (\ie$\bm{x}^0_i$) and $\bm{s}^t_i$ (\ie$\bm{s}^0_i$) come from the proposal detection step. $\bm{e}^{t}_i$ is  soft-weighted embedding of class label and $[,]$ is a concatenate operation. $F_s$ and $F_e$ are learnable functions\footnote{For conciseness, we leave the details in supplementary material. \label{supple}}. All internal states $\{\bm{h}^t_i\}$ are fed into the following message module to compose communication messages among agents.

\textbf{\texttt{Message} Module.} Considering the communication between agent $i$ and $j$, the message module will compose message $M^t_{ij}$ and $M^t_{ji}$ for each agent. Specifically, the messages $M^t_{ij}$ for agent $i$ is a tuple $M^t_{ij} = (\bm{m}^t_j, \bm{m}^t_{ij})$ including:
\begin{equation}
\small
\begin{aligned}
    \bm{m}^t_j = F_{m1}(\bm{h}^t_j), \; \bm{m}^t_{ij} = F_{m2}(\bm{h}^t_{ij}),
\end{aligned}
\end{equation}
where $\bm{m}^t_j$ is a unary message which captures the identity of agent $j$ (\eg, the local object content), and $\bm{m}^t_{ij}$ is a pairwise message which models the interaction between two agents (\eg, the relative spatial layout). $\bm{h}^t_{ij}$ is the pairwise feature between agent $i$ and $j$, and its initialization is the union box feature extracted by the object detector. $F_{m*}$ are message composition functions\footref{supple}. All communication message between agent $i$ and the others (\ie, $\{M^t_{i*}\}$) and its internal state $\bm{h}^t_i$ are fed into the following update module to update the time-step feature for next round agent communication\footnote{
We dubbed the communication step as agent communication instead of message passing~\cite{xu2017scene, li2017scene} for two reasons: 1) To be consistent with the concept of the multi-agent framework, where agent communication represents passing message among agents. 2) To highlight the difference with existing message passing methods that our communication model disentangles the relationship prediction from the visual context modeling.
}.

\textbf{\texttt{Update} Module.} At each round communication, we use a soft-attention~\cite{chen2017sca} to fuse message from other agents:
\begin{equation}
\small
\begin{aligned}
    \alpha^t_j &= F_{att1}(\bm{h}^t_i, \bm{h}^t_j), \;
    \alpha^t_{ij} = F_{att2}(\bm{h}^t_i, \bm{h}^t_{ij}), \\
    \bm{x}^{t+1}_i & = F_{u1}(\bm{h}^t_i, \{\alpha^t_j\bm{m}^t_j\}, \{\alpha^t_{ij}\bm{m}^t_{ij}\}), \\
    \bm{h}^{t+1}_{ij} &= F_{u2}(\bm{h}^t_{ij}, \bm{h}^t_i, \bm{h}^t_j),
\end{aligned}
\end{equation}
where $\alpha^t_j$ and $\alpha^t_{ij}$ are attention weights to fuse different message, $F_{att*}$ and $F_{u*}$ are attention and update functions\footref{supple}.


\subsubsection{Visual Relationship Detection}
\begin{small}
$$ \textbf{Input}: \{(\bm{s}^T_i, \bm{h}^T_i)\} \longmapsto \textbf{Output}: \{r_{ij}\} $$
\end{small}
After $T$-round agent communication, all agents finish their states update. In inference stage, we greedily select the object labels $v^T_i$ based on confidence $\bm{s}^T_i$. Then the relation model predict the relationship class for any object pairs:
\begin{equation}
\small
\begin{aligned}
    r_{ij} = F_r(\bm{h}^T_i, \bm{h}^T_j, v^T_i, v^T_j),
\end{aligned}
\end{equation}
where $F_r$ is the relationship function\footref{supple}. After predicting relationship for all object pairs, we finally obtain the generated scene graph: ($\{v^T_i\}, \{r_{ij} \}$).

\subsection{Counterfactual Critic Multi-Agent Training} \label{sec:3.2}
We demonstrate the details of the training objective of CMAT, including: 1) multi-agent policy gradient for the graph-coherent objective, and 2) counterfactual critic for the local-sensitive objective. The dataflow of our CMAT in training stage is shown in Figure~\ref{fig:2} (~\trainpath~ path).

\subsubsection{Graph-Coherent Training Objective}
Almost all prior SGG works minimize the XE loss as the training objective. Given a generated scene graph $(\hat{\mathcal{V}}, \hat{\mathcal{E}})$ and its ground-truth $(\mathcal{V}^{gt}, \mathcal{E}^{gt})$, the objective is:
\begin{equation} \label{eq:5}
\small
\begin{aligned}
   L(\theta) =  \textstyle{\sum_{ij}} \left(\text{XE}(\hat{v}_i, v^{gt}_i) + \text{XE}(\hat{r}_{ij}, r^{gt}_{ij}) \right). 
\end{aligned}
\end{equation}
As can be seen in Eq.~\eqref{eq:5}, the XE based objective is essentially independent and penalizes errors at all nodes equally.

To address this problem, we propose to replace XE with the following two graph-level metrics for graph-coherent training objective of SGG: 1) \textbf{Recall@K}~\cite{lu2016visual}: It computes the fraction of the correct predicted triplets in the top $K$ confident predictions. 2) \textbf{SPICE}~\cite{anderson2016spice}: It is the F-score of predicted triplets precision and triplets recall. Being different from the XE loss, both Recall@K and SPICE are non-differentiable. Thus, our CMAT resorts to using the multi-agent policy gradient to optimize these objectives.

\subsubsection{Multi-Agent Policy Gradient}
We first describe formally the \emph{action}, \emph{policy} and \emph{state} in CMAT, then derive the expression of parameter gradients.

\textbf{Action.} 
The action space for each agent is the set of all possible object classes, \ie, $v^t_i$ is the action of agent $i$. We denote $V^t = \{v^t_i\}$ as the set of actions of all agents.

\textbf{State.}
We follow previous work~\cite{hausknecht2015deep} to use an LSTM (\texttt{extract} module) to encode the \textit{history} of each agent. The hidden state $\bm{h}^t_i$ can be regarded as an approximation of the partially-observable environment state for agent $i$. We denote $H^t = \{\bm{h}^t_i\} $ as the set of states of all agents.

\textbf{Policy.}
The stochastic policy for each agent is the object classifier. In the training stage, the action is sampled based on the object class distribution, \ie, $\bm{p}^T_i = \text{softmax}(\bm{s}^T_i)$.

Because our CMAT only samples actions for each agents after $T$-round agent communication, based on the policy gradient theorem~\cite{sutton2000policy}, the (stochastic) gradient for the cooperative multi-agent in CMAT is:
\begin{align}
\begin{small}
\nabla_{\theta} J \approx \sum^n_{i=1} \nabla_{\theta} \log \bm{p}^T_i (v^T_i|h^T_i; \theta)Q(H^T, V^T),
\end{small}
\end{align}
where $Q(H^T, V^T)$ is the state-action value function. Instead of learning an independent network to fit the function $Q$ and approximate reward like actor-critic works~\cite{bahdanau2017actor, lowe2017multi, konda2000actor}; in our CMAT, we follow~\cite{rao2018learning} to directly use the real global reward to replace $Q$. The reasons are as follows: 1) The number of agents and possible actions for each agent in SGG are much larger than the previous multi-agent policy gradient settings, thus the number of training samples is insufficient to train an accurate value function. 2) This can reduce the model complexity and speed up the training procedures. Thus, the gradient for our CMAT becomes:
\begin{align} \label{eq:8}
\begin{small}
\nabla_{\theta} J \approx \sum^n_{i=1} \nabla_{\theta} \log \bm{p}^t_i (v^T_i|h^T_i; \theta) R(H^T, V^T),
\end{small}
\end{align}
where $R(H^T, V^T)$ is the real graph-level reward (\ie, Recall@K or SPICE). It is worth noting that the reward $R(H^T, V^T)$ is a learnable reward function which includes a relation detection model.

\subsubsection{Local-Sensitive Training Objective}
\begin{figure}[htbp]
\vspace{-0.8em}
\begin{minipage}[c]{0.4\linewidth}
\includegraphics[width=1\linewidth]{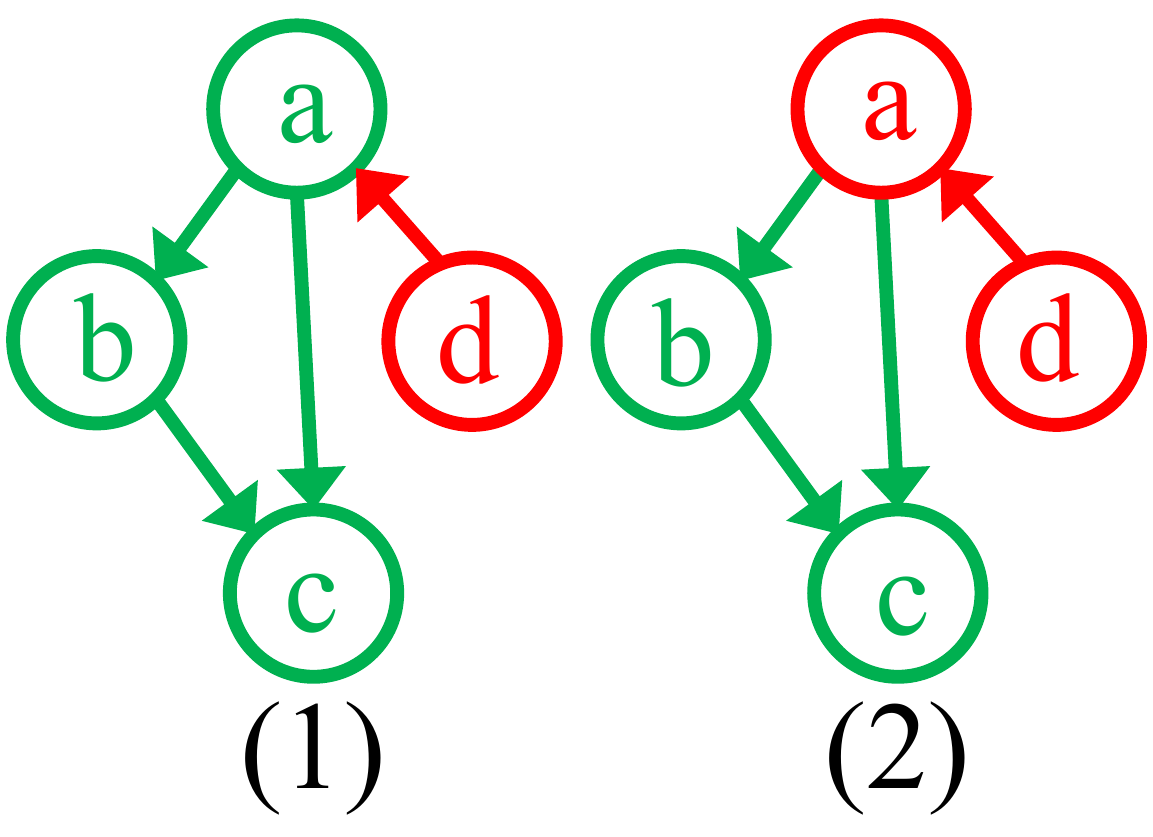}
\end{minipage}\hfill
\begin{minipage}[c]{0.55\linewidth}
\caption{\footnotesize (1)(2) are two generated scene graph results. The color green and red represents right and wrong prediction, respectively. The graph-level reward for this toy example is the number of right predicted triplets minus the number of wrong predicted triplets.}
\label{fig:local-sensitive}
\end{minipage}\vspace{-5mm}
\end{figure}

As can been seen in Eq.~\eqref{eq:8}, the graph-level reward can be considered as a global pooling contribution from all the local predictions, \ie, the reward for all the $n$ agents are identical. We demonstrate the negative impact of this situation with a toy example as shown in Figure~\ref{fig:local-sensitive}.

Suppose all predictions of two generated scene graph are identical, except that the prediction of node ``a" is different. Based on Eq.~\eqref{eq:8}, all nodes in the first graph and second graph get a positive reward (\ie, 3 (right) -1(wrong) = +2) and a negative reward (\ie, 1 (right)-3 (wrong) = -2), respectively. The predictions for the nodes ``b",``c", and ``d" are identical in the two graphs, but their gradient directions for optimization are totally different, which results in many inefficient optimization iteration steps. Thus, the training objective of SGG should be local-sensitive, \ie, it can identify the contribution of each local prediction to provide an efficient training signal for each agent.

\subsubsection{Counterfactual Critic}

An intuitive solution, for identifying the contribution of a specific agent's action, is to replace the default action of the target agent with other actions. Formally, $R(H^T, V^T) - R(H^T, (V^T_{-i}, \tilde{v}^T_i))$ can reflect the true influence of action $v^T_i$, where $V^T_{-i}$ represents all agents except agent $i$ (\ie, the others $n-1$ agents) using the default action, and agent $i$ takes a new action $\tilde{v}^T_i$.

\begin{figure}[h]
	\centering
	\includegraphics[width=1\linewidth]{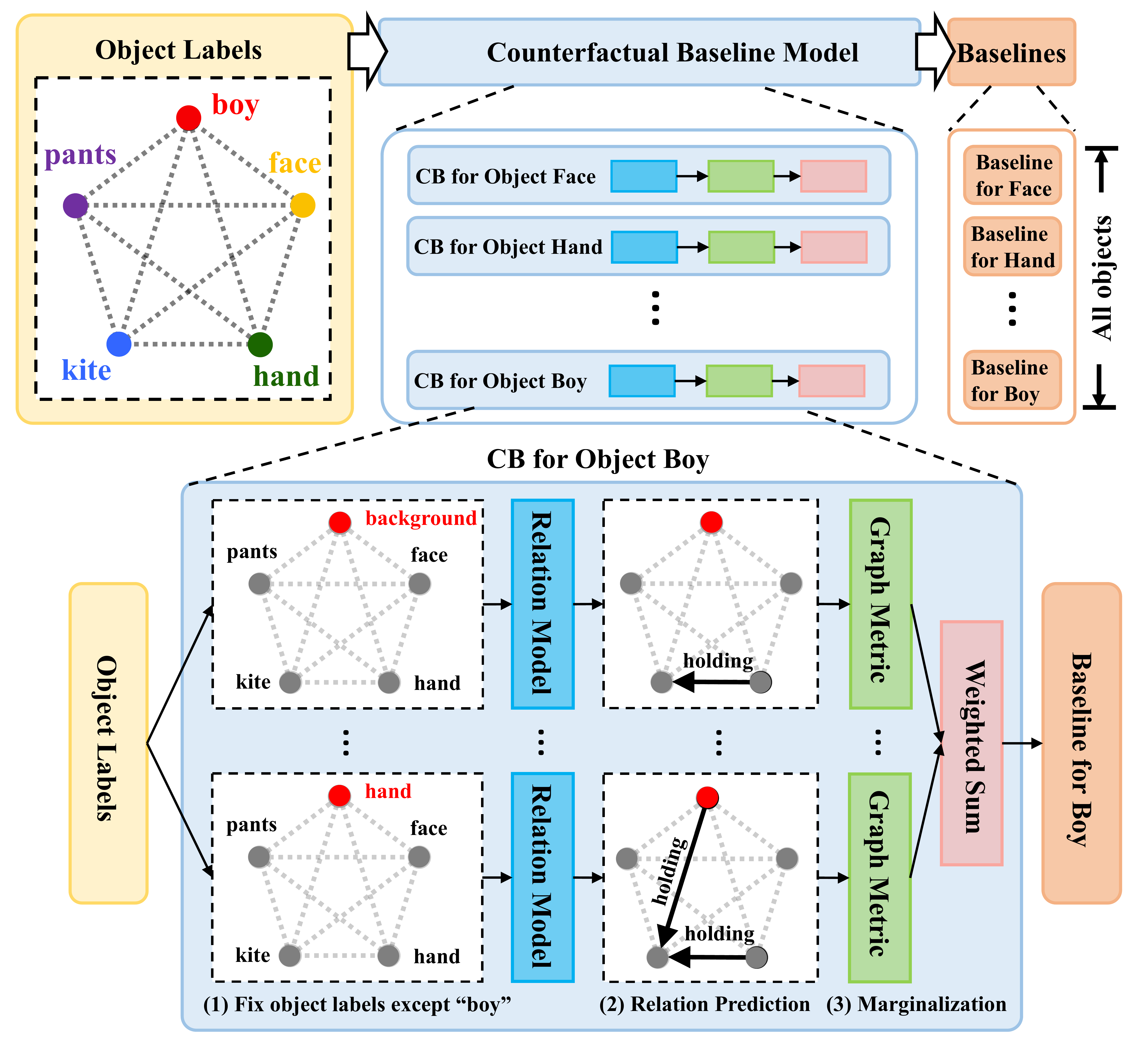}
	\caption{The illustration of the counterfactual baseline (CB) model in Figure~\ref{fig:2}. For this given image, the model calculates CB for all agents (\eg, \textcolor{red}{\texttt{boy}}, \textcolor[rgb]{0.99, 0.76, 0.00}{\texttt{face}}, \textcolor[rgb]{0.13, 0.55, 0.13}{\texttt{hand}}, \textcolor[rgb]{0.56, 0.0, 1.0}{\texttt{pants}}, and \textcolor[rgb]{0.0, 0.0, 1.0}{\texttt{kite}}). As shown in the bottom, for the CB for \textcolor{red}{\texttt{boy}}, we traverse to replace class label \texttt{boy} to all possible classes (\eg, \texttt{background} $,...,$ \texttt{hand} \etal) and marginalize these rewards.}
	\label{fig:4} \vspace{-0.5em}
\end{figure}

Since the new action $\tilde{v}^T_i$ for agent $i$ has $|\mathcal{C}|$ choices, and we can obtain totally different results for $R(H^T, (V^T_{-i}, \tilde{v}^T_i))$ with different action choices. To more precisely approximate the individual reward of the default action of agent $i$ (\ie, $v^T_i$), we marginalize the rewards when agent $i$ traverse all possible actions: $ \text{CB}^i(H^T, V^T) = \sum \bm{p}^T_i(\tilde{v}^T_i) R(H^T, (V^T_{-i}, \tilde{v}^T_i))$, where $\text{CB}^i(H^T, V^T)$ is the \textbf{\emph{counterfactual baseline}} for the action of agent $i$. The counterfactual baseline represents the average global-level reward that the model should receive when all other agents take default actions and regardless of the action of agent $i$. The illustration of counterfactual baseline model (Figure~\ref{fig:2}) in CMAT is shown in Figure~\ref{fig:4}.

Given the global reward $R(H^T, V^T)$ and counterfactual baseline $\text{CB}^i(H^T, V^T)$ for action $\bm{v}^T_i$ of agent $i$, the disentangled contribution of the action of agent $i$ is:
\begin{align}
\begin{small}
    A^i(H^T, V^T) = R(H^T, V^T) - \text{CB}^i(H^T, V^T).
\end{small}
\end{align}

Note that $A^i(H^T, V^T)$  can be considered as the \emph{advantage} in actor-critic methods~\cite{sutton1998reinforcement, mnih2016asynchronous}, $\text{CB}^i(H^T, V^T)$ can be regarded as a \emph{baseline} in policy gradient methods, which reduces the variance of gradient estimation. The whole network to calculate $A^i(H^T, V^T)$ is dubbed as the \textbf{\emph{counterfactual critic}}\footnote{
Although the critic in CMAT is not a value function to estimate the reward as in actor-critic, we dubbed it as critic for two reasons: 1) The essence of a critic is calculating advantages for the actions of policy network. As in previous policy gradient work~\cite{rennie2017self}, the critic can be an inference algorithm without a value function. 2) The critic in CMAT includes a learnable relation model, which will also update its parameters at training.
} (Figure~\ref{fig:2}). Then the gradient becomes:
\begin{align}
\begin{small}
\nabla_{\theta} J \approx \sum^n_{i=1} \nabla_{\theta} \log \bm{p}^T_i (v^T_i|h^T_i; \theta) A^i(H^T, V^T).
\end{small}
\end{align}

Finally, we incorporate the auxiliary XE supervised loss (weighted by a trade-off $\alpha$) for an end-to-end training, and the overall gradient is:
\begin{equation}
\small
\begin{split}
\nabla_{\theta} J \approx  \overbrace{\sum^n_{i=1} \nabla_{\theta} \log \bm{p}^T_i (v^T_i|h^T_i; \theta) A^i(H^T, V^T)}^{\text{CMAT}} + \\
\underbrace{\alpha\sum^n_{i=1}\sum^n_{j=1} \nabla_{\theta} \log \bm{p}_{ij}(r_{ij})}_{\text{XE for relationships}} + 
\underbrace{\alpha \sum^n_{i=1} \nabla_{\theta} \log \bm{p}^T_i(v^T_i)}_{\text{XE for objects}},
\end{split}
\end{equation}
where CMAT encourages visual context exploration and XE stabilizes the training~\cite{rao2018learning}. We also follow~\cite{xu2015show, hu2017learning} to add an entropy term to regularize $\{\bm{p}^T_i\}_i$.


\section{Experiments}

\noindent\textbf{Dataset.} We evaluated our method for SGG on the challenging benchmark: Visual Genome (VG)~\cite{krishna2017visual}. For fair comparisons, we used the released data preprocessing and splits which had been widely-used in~\cite{xu2017scene, zellers2018neural, newell2017pixels, yang2018graph, herzig2018mapping}. The release selects the most frequent 150 object categories and 50 predicate classes. After preprocessing, each image has 11.5 objects and 6.2 relationships on average. The released split uses 70\% of images for training (including 5K images as validation set) and 30\% of images for test. 

\noindent\textbf{Settings.} As the conventions in~\cite{xu2017scene, zellers2018neural, jae2018tensorize}, we evaluate SGG on three tasks: \textbf{Predicate Classification} (PredCls): Given the ground-truth object bounding boxes and class labels, we need to predict the visual relationship classes among all the object pairs. \textbf{Scene Graph Classification} (SGCls): Given the ground-truth object bounding boxes, we need to predict both the object and pairwise relationship classes. \textbf{Scene Graph Detection} (SGDet): Given an image, we need to detect the objects and predict their pairwise relationship classes. In particular, the object detection needs to localize both the subject and object with at least 0.5 IoU with the ground-truth. As the conventions in~\cite{zellers2018neural, jae2018tensorize}, we used Recall@20 (R@20), Recall@50 (R@50), and Recall@100 (R@100) as the evaluation metrics.

\begin{table*}
\small
\begin{center}
\scalebox{0.95}{
\begin{tabular}{l|l|ccc|ccc|ccc|c}
\hline
\multicolumn{1}{c|}{} & \multicolumn{1}{c|}{} & \multicolumn{3}{c|}{SGDet} & \multicolumn{3}{c|}{SGCls} & \multicolumn{3}{c|}{PredCls} \\
& Model & R@20 & R@50 & R@100  & R@20 & R@50 & R@100 & R@20 & R@50 & R@100 & Mean \\ 
\hline\hline
\multirow{10}{*}{Constraint} & VRD~\cite{lu2016visual} & - & 0.3 & 0.5 & - & 11.8 & 14.1 & - & 27.9 & 35.0 & 14.9 \\
& IMP~\cite{xu2017scene} & - & 3.4 & 4.2 & - & 21.7 & 24.4 & - & 44.8 & 53.0 & 25.3 \\
& MSDN~\cite{li2017scene, yang2018graph} & - & 7.0 & 9.1 & - & 27.6 & 29.9 & - & 53.2 & 57.9 & 30.8 \\
& AsscEmbed~\cite{newell2017pixels} & 6.5 & 8.1 & 8.2 & 18.2 & 21.8 & 22.6 & 47.9 & 54.1 & 55.4 & 28.3 \\
& FREQ+$^\diamond$~\cite{zellers2018neural} & 20.1 & 26.2 & 30.1 & 29.3 & 32.3 & 32.9 & 53.6 & 60.6 & 62.2 & 40.7 \\
& IMP+$^\diamond$~\cite{xu2017scene, zellers2018neural} & 14.6 & 20.7 & 24.5 & 31.7 & 34.6 & 35.4 & 52.7 & 59.3 & 61.3 & 39.3 \\
& TFR~\cite{jae2018tensorize} & 3.4 & 4.8 & 6.0 & 19.6 & 24.3 & 26.6 & 40.1 & 51.9 & 58.3 & 28.7 \\
& MOTIFS$^\diamond$~\cite{zellers2018neural} & 21.4 & 27.2 & 30.3 & 32.9 & 35.8 & 36.5 & 58.5 & 65.2 & 67.1 & 43.7 \\
& Graph-RCNN~\cite{yang2018graph} & - & 11.4 & 13.7 & - & 29.6 & 31.6 & - & 54.2 & 59.1 & 33.2 \\
& GPI$^\diamond$~\cite{herzig2018mapping} & - & - & - & - & 36.5 & 38.8 & - & 65.1 & 66.9 & -  \\
& KER$^\diamond$~\cite{chen2019knowledge} & - & 27.1 & 29.8 & - & 36.7 & 37.4 & - & 65.8 & 67.6 & 44.1 \\
& \textbf{CMAT} & \textbf{22.1} & \textbf{27.9} & \textbf{31.2} & \textbf{35.9} & \textbf{39.0} & \textbf{39.8} & \textbf{60.2} & \textbf{66.4} & \textbf{68.1} & \textbf{45.4} \\
\hline
\multirow{5}{*}{No constraint} & AsscEmbed~\cite{newell2017pixels} & - & 9.7 & 11.3 & - & 26.5 & 30.0 & - & 68.0 & 75.2 & 36.8 \\
& IMP+$^\diamond$~\cite{xu2017scene, zellers2018neural} & - & 22.0 & 27.4 & - & 43.4 & 47.2 & - & 75.2 & 83.6 & 49.8 \\
& FREQ+$^\diamond$~\cite{zellers2018neural} & - & 28.6 & 34.4 & - & 39.0 & 43.4 & - & 75.7 & 82.9 & 50.6 \\
& MOTIFS$^\diamond$~\cite{zellers2018neural} & 22.8 & 30.5 & 35.8 & 37.6 & 44.5 & 47.7 & 66.6 & 81.1 & 88.3 & 54.7 \\
& KER$^\diamond$~\cite{chen2019knowledge} & - & 30.9 & 35.8 & - & 45.9 & 49.0 & - & 81.9 & 88.9 & 55.4 \\
& \textbf{CMAT} & \textbf{23.7} & \textbf{31.6} & \textbf{36.8} & \textbf{41.0} & \textbf{48.6} & \textbf{52.0} & \textbf{68.9} & \textbf{83.2} & \textbf{90.1} & \textbf{57.0} \\
\hline
\end{tabular}
} 
\end{center}
\vspace{-1.5em}
\caption{Performance (\%) compared with the state-of-the-art methods w/o graph constraint on VG~\cite{krishna2017visual}. Since some works doesn't evaluate on R@20, we compute the mean on all tasks over R@50 and R@100. $^\diamond$ denotes the methods using the same object detector as ours.}
\label{tab:1} \vspace{-1.0em}
\end{table*}

\begin{table*}
\subfloat[Results (\%) of different reward choices.]{
\tablestyle{2.5pt}{1.05}\begin{tabular}{c|x{22}x{22}x{22}x{22}}
\hline
& & XE & R@20 & SPICE  \\
\hline
\multirow{2}{*}{SGCls} & R@20 & 34.08 & \textbf{35.93} & 35.27  \\
& SPICE & 15.39 & \textbf{16.01} & 15.90 \\
\hline
\multirow{2}{*}{SGDet} & R@20 & 16.23 & \textbf{16.53} & 16.51  \\
& SPICE & 7.48 & \textbf{7.66} & 7.64  \\
\hline\hline
\end{tabular}}\hspace{3mm}
\subfloat[Results (\%) of different baseline types.]{
\tablestyle{2.5pt}{1.05}\begin{tabular}{c|x{22}x{22}x{22}x{22}x{22}}
\hline
& & XE & MA & SC & CF  \\
\hline
\multirow{3}{*}{SGCls} & R@20 & 34.08 & 34.76 & 34.68 & \textbf{35.93} \\
& R@50 & 36.90 & 37.58 & 37.54 & \textbf{39.00} \\
& R@100 & 37.61 & 38.29 & 38.25 & \textbf{39.75} \\
\hline
\multirow{3}{*}{SGDet} & R@20 & 16.23 & 16.07 & 16.37 & \textbf{16.53} \\
& R@50 & 20.62 & 20.41 & 20.82 & \textbf{20.95} \\
& R@100 & 23.24 & 23.02 & 23.41 & \textbf{23.62} \\
\hline\hline
\end{tabular}}\hspace{3mm}
\subfloat[Results (\%) of different \#communication steps.]{
\tablestyle{2.5pt}{1.05}\begin{tabular}{c|x{22}x{22}x{22}x{22}x{22}}
\hline
& & 2-step & 3-step & 4-step & 5-step  \\
\hline
\multirow{3}{*}{SGCls} & R@20 & 35.09 & 35.25 & 35.40 & \textbf{35.93} \\
& R@50 & 37.95  & 38.19 & 38.37 & \textbf{39.00} \\
& R@100 & 38.67 & 38.91 & 39.09 & \textbf{39.75} \\
\hline
\multirow{3}{*}{SGDet} & R@20 & 16.35 & 16.43 & 16.47 & \textbf{16.53} \\
& R@50 & 20.89 & 20.88 & 20.92 & \textbf{20.95} \\
& R@100 & 23.49 & 23.50 & 23.54 & \textbf{23.62} \\
\hline\hline
\end{tabular}}\hspace{3mm}
\vspace{-0.5em}
\caption{\textbf{Ablations}. All results are with graph constraints. XE: The initialization performance after supervised XE pre-training. For clarity, the results of SGDet without post-processing are shown.}
\vspace{-0.5em}
\label{tab:2}
\end{table*}

\subsection{Implementation Details}
\noindent\textbf{Object Detector.} For fair comparisons with previous works, we adopted the same object detector as~\cite{zellers2018neural}. Specifically, the object detector is a Faster-RCNN~\cite{ren2015faster} with VGG backbone~\cite{simonyan2015very}. Moreover, the anchor box size and aspect ratio are adjusted similar to YOLO-9000~\cite{redmon2017yolo9000}, and the RoIPooling layer is replaced with the RoIAlign layer~\cite{he2017mask}.

\noindent\textbf{Training Details.}
Following the previous policy gradient works that use a supervised pre-training step as model initialization (\emph{aka}, teacher forcing), our CMAT also utilized this two-stage training strategy. In the supervised training stage, we froze the layers before the ROIAlign layer and optimized the whole framework with the sum of objects and relationships XE losses. The batch size and initial learning rate were set to 6 and $10^{-3}$, respectively. In the policy gradient training stage, the initial learning rate is set to $3\times10^{-5}$. For SGDet, since the number of all possible relationship pairs are huge (\eg, 64 objects leads to $\approx$ 4,000 pairs), we followed~\cite{zellers2018neural} that only considers the relationships between two objects with overlapped bounding boxes, which reduced the number of object pairs to around 1,000.

\noindent\textbf{Speed vs. Accuracy Trade-off.}
In the policy gradient training stage, the complete counterfactual critic calculation needs to sum over all possible object classes, which is significantly time-consuming (over 9,600 ($\approx 151\times 64$) times graph-level evaluation at each iteration). Fortunately, we noticed that only a few classes for each agent have large prediction confidence. To make a trade-off between training speed and accuracy, we only sum over the two highest positive classes and the background class probabilities to estimate the counterfactual baseline. In our experiments, this approximation only results in a slight performance drop but 70x faster training time.

\noindent\textbf{Post-processing for SGDet.} For SGDet, we followed the post-processing step in~\cite{zellers2018neural, Zhang_2019_CVPR} for a fair comparison. After predicting the object class probabilities for each RoI, we used a per-class NMS to select the RoI class and its corresponding class-specific offsets from Faster-RCNN. The IoU threshold in NMS was set to 0.5 in our experiments.

\begin{figure*}[htbp]
	\centering
	\includegraphics[width=1\linewidth]{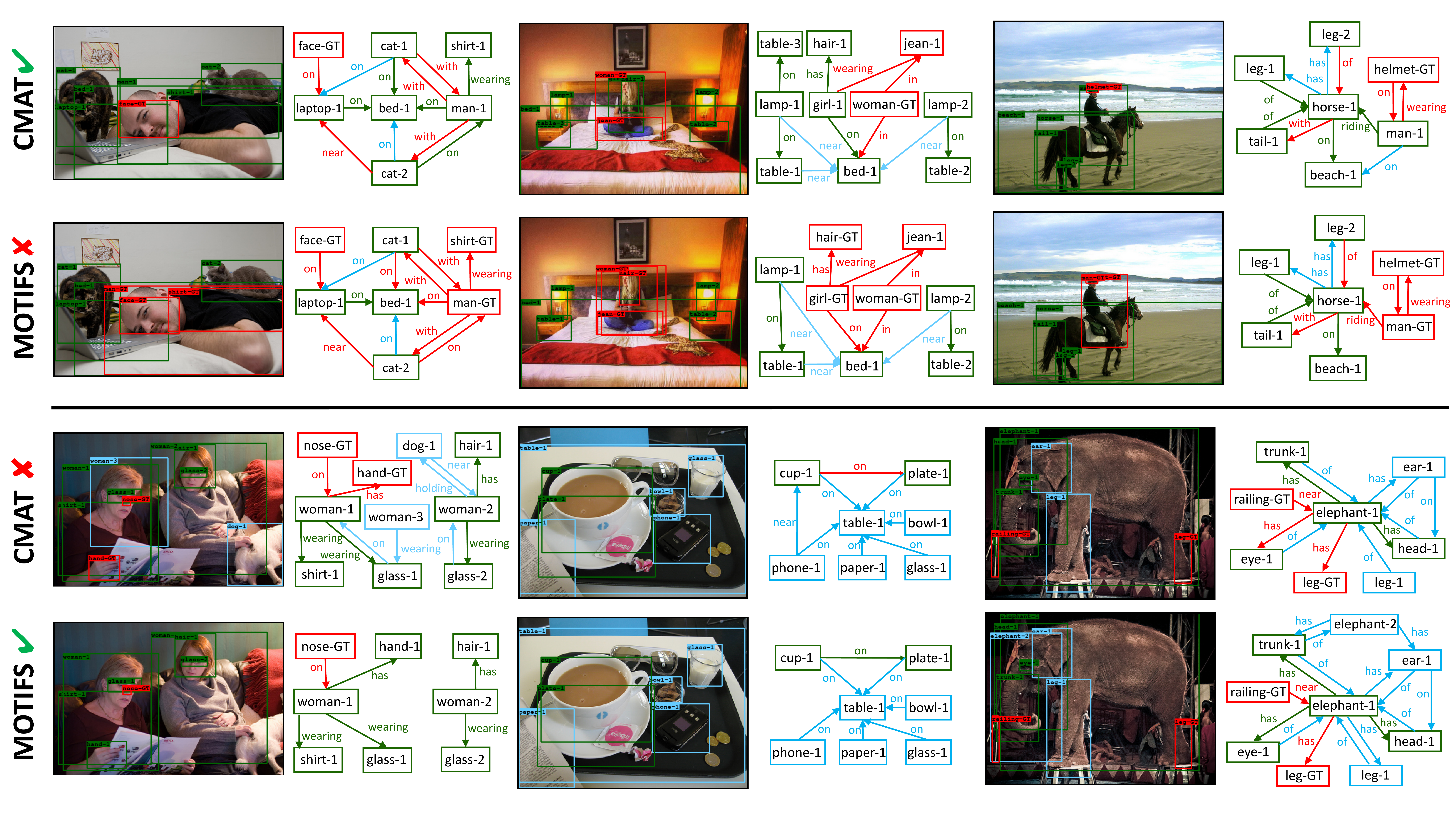}
	\vspace{-2.0em}
	\caption{Qualitative results showing comparisons between higher R@20 (green tick) and lower R@20 (red cross) by CMAT and MOTIFS in SGDet. Green boxes are detected boxes with IoU large 0.5 with the ground-truth, blue boxes are detected but not labeled, red boxes are ground-truth with no match. Green edges are true positive predicted by each model at the R@20 setting, red edges are false negatives, and blue edges are false positives. Only detected boxes overlapped with ground-truth are shown.}
	\vspace{-1.0em}
\label{fig:5}
\end{figure*}

\subsection{Ablative Studies}
We run a number of ablations to analyze CMAT, including the graph-level reward choice (for graph-coherent characteristic), the effectiveness of counterfactual baseline (for local-sensitive characteristic), and the early saturation problem in agent communication model. Results are shown in Table~\ref{tab:2} and discussed in detail next.

\noindent\textbf{Graph-level Reward Choices.}
To investigate the influence of choosing different graph-level metrics as the training reward, we compared two metrics: \textbf{Recall@K} and \textbf{SPICE}. In particular, we used the top-20 confident triplets as the predictions to calculate Recall and SPICE. The results are shown in Table~\ref{tab:2} (a). We can observe that using both Recall and SPICE as the training reward can consistently improve the XE pre-trained model, because \textbf{the graph-level metrics is a graph-coherent objective}. Meanwhile, using Recall@K as training reward can always get slightly better performance than SPICE, because SPICE is not a suitable evaluation metric for the incomplete annotation nature of VG. Therefore, we used Recall@K as our training reward in the rest of the experiments.

\noindent\textbf{Policy Gradient Baselines.}
To evaluate the effectiveness of our counterfactual  baseline (\textbf{CF}), we compared it with other two widely-used baselines in policy gradient: Moving Average (\textbf{MA})~\cite{weaver2001optimal} and Self-Critical (\textbf{SC})~\cite{rennie2017self}. MA is a moving average constant over the recent rewards~\cite{xu2015show, hu2017learning}. SC is the received reward when model directly takes greedy actions as in the test. From Table~\ref{tab:2} (b), we can observe that our CF baseline consistently improves the supervised initialization and outperforms others. Meanwhile, MA and SC can only improve the performance slightly or even worsen it. Because \textbf{the CF baseline is a local-sensitive objective} and provides a more effective training signal for each agent, while MA and SC baselines are only globally pooling rewards which are still not local-sensitive.

\noindent\textbf{\# of Communication Steps.} 
To investigate the early saturation issue in message passing models~\cite{xu2017scene, li2017scene}, we compared the performance of CMAT with different numbers of communication steps from 2 to 5. From Table~\ref{tab:2} (c), we can observe the trend seems contiguously better with the increase of communication step. Due to the GPU memory limit, we conducted experiments up to 5 steps. Compared to existing message passing methods, the reason why CMAT can avoid the early saturation issue is that our agent communication model discards the widely-used relationship nodes.

\subsection{Comparisons with State-of-the-Arts}

\noindent\textbf{Settings.}
We compared CMAT with the state-of-the-art models. According to whether the model encodes context, we group these methods into: 1) \textbf{VRD}~\cite{lu2016visual}, \textbf{AsscEmbed}~\cite{newell2017pixels}, \textbf{FREQ}~\cite{zellers2018neural} are independent inference models, which predict object and relation classes independently. 2) \textbf{MSDN}~\cite{li2017scene}, \textbf{IMP}~\cite{xu2017scene}, \textbf{TFR}~\cite{jae2018tensorize}, \textbf{MOTIFS}~\cite{zellers2018neural}, \textbf{Graph-RCNN}~\cite{yang2018graph}, \textbf{GPI}~\cite{herzig2018mapping}, \textbf{KER}~\cite{chen2019knowledge} are joint inference models, which adopt message passing to encode the context. All these models are optimized by XE based training objective. 

\noindent\textbf{Quantitative Results.} 
The quantitative results are reported in Table~\ref{tab:1}. From Table~\ref{tab:1}, we can observe that our CMAT model achieves the state-of-the-art performance under all evaluation metrics. It is worth noting that CMAT can especially improve the performance of SGCls significantly (\ie, 3.4\% and 4.3\% absolute improvement in with and without graph constraint setting respectively), which means our CMAT model can substantially improve the object label predictions compared to others. The improvements in object label predictions meet our CMAT design, where the action of each agent is to predict an object label. Meanwhile, it also demonstrates the effectiveness of counterfactual critic multi-agent training for message passing models (agent communication) compared with XE based training. For PredCls task, even we use the easiest visual relationship model and it achieves the best performance, which means the input for relationship model (\ie, the state of agent) can better capture the internal state of each agent. Meanwhile, it is worth noting that any stronger relationship model can seamlessly be incorporated into our CMAT. For SGDet task, the improvements are not as significant as the SGCls, the reason may come from the imperfect and noisy detected bounding boxes. 

\noindent\textbf{Qualitative Results.}
Figure~\ref{fig:5} shows the qualitative results compared with MOTIFS. From the results in the top two rows, we can observe that CMAT rarely mistakes at the important hub nodes such as the \texttt{man} or \texttt{girl}, because CMAT directly optimizes the graph-coherent objective. From the results in the bottom two rows, the mistakes of CMAT always come from the incomplete annotation of VG: CMAT can detect more false positive (the blue color) objects and relationship than MOTIFS. Since the evaluation metric (\ie, Recall@K) is based on the ranking of labeled triplet confidence, thus, detecting more reasonable false positive results with high confidence will worsen the results.

\section{Conclusions}
We proposed a novel method CMAT to address the inherent problem with XE based objective in SGG: it is not a graph-coherent objective. CMAT solves the problems by 1) formulating SGG as a multi-agent cooperative task, and using graph-level metrics as the training reward. 2) disentangling the individual contribution of each object to allow a more focused training signal. We validated the effectiveness of CMAT through extensive comparative and ablative experiments. Moving forward, we are going to 1) design a more effective graph-level metric to guide the CMAT training and 2) apply CMAT in downstream tasks such as VQA~\cite{xu2017video, ye2017video}, dialog~\cite{niu2019recursive}, and captioning~\cite{xu2018dual}.

\small \noindent\textbf{Acknowledgement} This work is supported by Zhejiang Natural Science Foundation (LR19F020002, LZ17F020001), National Natural Science Foundation of China (6197020369, 61572431), National Key Research and Development Program of China (SQ2018AAA010010), the Fundamental Research Funds for the Central Universities and Chinese Knowledge Center for Engineering Sciences and Technology. L. Chen is supported by 2018 Zhejiang University Academic Award for Outstanding Doctoral Candidates. H. Zhang is supported by NTU Data Science and Artificial Intelligence Research Center (DSAIR) and NTU-Alibaba Lab.

{\small
\bibliographystyle{ieee_fullname}
\bibliography{egbib}

\begin{thebibliography}{10}\itemsep=-1pt

\bibitem{anderson2016spice}
Peter Anderson, Basura Fernando, Mark Johnson, and Stephen Gould.
\newblock Spice: Semantic propositional image caption evaluation.
\newblock In {\em ECCV}, 2016.

\bibitem{bahdanau2017actor}
Dzmitry Bahdanau, Philemon Brakel, Kelvin Xu, Anirudh Goyal, Ryan Lowe, Joelle
  Pineau, Aaron Courville, and Yoshua Bengio.
\newblock An actor-critic algorithm for sequence prediction.
\newblock In {\em ICLR}, 2017.

\bibitem{caicedo2015active}
Juan~C Caicedo and Svetlana Lazebnik.
\newblock Active object localization with deep reinforcement learning.
\newblock In {\em ICCV}, 2015.

\bibitem{chen2017query}
Kan Chen, Rama Kovvuri, and Ram Nevatia.
\newblock Query-guided regression network with context policy for phrase
  grounding.
\newblock In {\em ICCV}, 2017.

\bibitem{chen2017sca}
Long Chen, Hanwang Zhang, Jun Xiao, Liqiang Nie, Jian Shao, Wei Liu, and
  Tat-Seng Chua.
\newblock Sca-cnn: Spatial and channel-wise attention in convolutional networks
  for image captioning.
\newblock In {\em CVPR}, 2017.

\bibitem{chen2019knowledge}
Tianshui Chen, Weihao Yu, Riquan Chen, and Liang Lin.
\newblock Knowledge-embedded routing network for scene graph generation.
\newblock In {\em CVPR}, 2019.

\bibitem{dai2017detecting}
Bo Dai, Yuqi Zhang, and Dahua Lin.
\newblock Detecting visual relationships with deep relational networks.
\newblock In {\em CVPR}, 2017.

\bibitem{das2017learning}
Abhishek Das, Satwik Kottur, Jos{\'e}~MF Moura, Stefan Lee, and Dhruv Batra.
\newblock Learning cooperative visual dialog agents with deep reinforcement
  learning.
\newblock In {\em ICCV}, 2017.

\bibitem{divvala2009empirical}
Santosh~K Divvala, Derek Hoiem, James~H Hays, Alexei~A Efros, and Martial
  Hebert.
\newblock An empirical study of context in object detection.
\newblock In {\em CVPR}, 2009.

\bibitem{foerster2016learning}
Jakob Foerster, Ioannis~Alexandros Assael, Nando de Freitas, and Shimon
  Whiteson.
\newblock Learning to communicate with deep multi-agent reinforcement learning.
\newblock In {\em NeurIPS}, 2016.

\bibitem{foerster2018counterfactual}
Jakob Foerster, Gregory Farquhar, Triantafyllos Afouras, Nantas Nardelli, and
  Shimon Whiteson.
\newblock Counterfactual multi-agent policy gradients.
\newblock In {\em AAAI}, 2018.

\bibitem{foerster2017stabilising}
Jakob Foerster, Nantas Nardelli, Gregory Farquhar, Triantafyllos Afouras,
  Philip~HS Torr, Pushmeet Kohli, and Shimon Whiteson.
\newblock Stabilising experience replay for deep multi-agent reinforcement
  learning.
\newblock In {\em ICML}, 2017.

\bibitem{gu2019scene}
Jiuxiang Gu, Handong Zhao, Zhe Lin, Sheng Li, Jianfei Cai, and Mingyang Ling.
\newblock Scene graph generation with external knowledge and image
  reconstruction.
\newblock In {\em CVPR}, 2019.

\bibitem{Haurilet_2019_CVPR}
Monica Haurilet, Alina Roitberg, and Rainer Stiefelhagen.
\newblock It's not about the journey; it's about the destination: Following
  soft paths under question-guidance for visual reasoning.
\newblock In {\em CVPR}, 2019.

\bibitem{hausknecht2015deep}
Matthew Hausknecht and Peter Stone.
\newblock Deep recurrent q-learning for partially observable mdps.
\newblock In {\em AAAI}, 2015.

\bibitem{he2017mask}
Kaiming He, Georgia Gkioxari, Piotr Doll{\'a}r, and Ross Girshick.
\newblock Mask r-cnn.
\newblock In {\em ICCV}, 2017.

\bibitem{herzig2018mapping}
Roei Herzig, Moshiko Raboh, Gal Chechik, Jonathan Berant, and Amir Globerson.
\newblock Mapping images to scene graphs with permutation-invariant structured
  prediction.
\newblock In {\em NeurIPS}, 2018.

\bibitem{hu2017learning}
Ronghang Hu, Jacob Andreas, Marcus Rohrbach, Trevor Darrell, and Kate Saenko.
\newblock Learning to reason: End-to-end module networks for visual question
  answering.
\newblock In {\em ICCV}, 2017.

\bibitem{hudson2019gqa}
Drew~A Hudson and Christopher~D Manning.
\newblock Gqa: A new dataset for real-world visual reasoning and compositional
  question answering.
\newblock In {\em CVPR}, 2019.

\bibitem{jae2018tensorize}
Seong Jae~Hwang, Sathya~N Ravi, Zirui Tao, Hyunwoo~J Kim, Maxwell~D Collins,
  and Vikas Singh.
\newblock Tensorize, factorize and regularize: Robust visual relationship
  learning.
\newblock In {\em CVPR}, 2018.

\bibitem{jie2016tree}
Zequn Jie, Xiaodan Liang, Jiashi Feng, Xiaojie Jin, Wen Lu, and Shuicheng Yan.
\newblock Tree-structured reinforcement learning for sequential object
  localization.
\newblock In {\em NeurIPS}, 2016.

\bibitem{johnson2017inferring}
Justin Johnson, Bharath Hariharan, Laurens van~der Maaten, Judy Hoffman, Li
  Fei-Fei, C~Lawrence Zitnick, and Ross~B Girshick.
\newblock Inferring and executing programs for visual reasoning.
\newblock In {\em ICCV}, 2017.

\bibitem{johnson2015image}
Justin Johnson, Ranjay Krishna, Michael Stark, Li-Jia Li, David Shamma, Michael
  Bernstein, and Li Fei-Fei.
\newblock Image retrieval using scene graphs.
\newblock In {\em CVPR}, 2015.

\bibitem{Kim_2019_CVPR}
Dong-Jin Kim, Jinsoo Choi, Tae-Hyun Oh, and In~So Kweon.
\newblock Dense relational captioning: Triple-stream networks for
  relationship-based captioning.
\newblock In {\em CVPR}, 2019.

\bibitem{konda2000actor}
Vijay~R Konda and John~N Tsitsiklis.
\newblock Actor-critic algorithms.
\newblock In {\em NeurIPS}, 2000.

\bibitem{krahenbuhl2011efficient}
Philipp Kr{\"a}henb{\"u}hl and Vladlen Koltun.
\newblock Efficient inference in fully connected crfs with gaussian edge
  potentials.
\newblock In {\em NeurIPS}, 2011.

\bibitem{krishna2017visual}
Ranjay Krishna, Yuke Zhu, Oliver Groth, Justin Johnson, Kenji Hata, Joshua
  Kravitz, Stephanie Chen, Yannis Kalantidis, Li-Jia Li, David~A Shamma, et~al.
\newblock Visual genome: Connecting language and vision using crowdsourced
  dense image annotations.
\newblock {\em IJCV}, 2017.

\bibitem{li2017vip}
Yikang Li, Wanli Ouyang, Xiaogang Wang, and Xiao'ou Tang.
\newblock Vip-cnn: Visual phrase guided convolutional neural network.
\newblock In {\em CVPR}, 2017.

\bibitem{li2018factorizable}
Yikang Li, Wanli Ouyang, Bolei Zhou, Yawen Cui, Jianping Shi, and Xiaogang
  Wang.
\newblock Factorizable net: An efficient subgraph-based framework for scene
  graph generation.
\newblock In {\em ECCV}, 2018.

\bibitem{li2017scene}
Yikang Li, Wanli Ouyang, Bolei Zhou, Kun Wang, and Xiaogang Wang.
\newblock Scene graph generation from objects, phrases and region captions.
\newblock In {\em ICCV}, 2017.

\bibitem{liang2017deep}
Xiaodan Liang, Lisa Lee, and Eric~P Xing.
\newblock Deep variation-structured reinforcement learning for visual
  relationship and attribute detection.
\newblock In {\em CVPR}, 2017.

\bibitem{liu2018context}
Daqing Liu, Zheng-Jun Zha, Hanwang Zhang, Yongdong Zhang, and Feng Wu.
\newblock Context-aware visual policy network for sequence-level image
  captioning.
\newblock In {\em ACM MM}, 2018.

\bibitem{liu2017improved}
Siqi Liu, Zhenhai Zhu, Ning Ye, Sergio Guadarrama, and Kevin Murphy.
\newblock Improved image captioning via policy gradient optimization of spider.
\newblock In {\em ICCV}, 2017.

\bibitem{liu2016ssd}
Wei Liu, Dragomir Anguelov, Dumitru Erhan, Christian Szegedy, Scott Reed,
  Cheng-Yang Fu, and Alexander~C Berg.
\newblock Ssd: Single shot multibox detector.
\newblock In {\em ECCV}, 2016.

\bibitem{long2015fully}
Jonathan Long, Evan Shelhamer, and Trevor Darrell.
\newblock Fully convolutional networks for semantic segmentation.
\newblock In {\em CVPR}, 2015.

\bibitem{lowe2017multi}
Ryan Lowe, Yi Wu, Aviv Tamar, Jean Harb, OpenAI~Pieter Abbeel, and Igor
  Mordatch.
\newblock Multi-agent actor-critic for mixed cooperative-competitive
  environments.
\newblock In {\em NeurIPS}, 2017.

\bibitem{lu2016visual}
Cewu Lu, Ranjay Krishna, Michael Bernstein, and Li Fei-Fei.
\newblock Visual relationship detection with language priors.
\newblock In {\em ECCV}, 2016.

\bibitem{mathe2016reinforcement}
Stefan Mathe, Aleksis Pirinen, and Cristian Sminchisescu.
\newblock Reinforcement learning for visual object detection.
\newblock In {\em CVPR}, 2016.

\bibitem{mnih2016asynchronous}
Volodymyr Mnih, Adria~Puigdomenech Badia, Mehdi Mirza, Alex Graves, Timothy
  Lillicrap, Tim Harley, David Silver, and Koray Kavukcuoglu.
\newblock Asynchronous methods for deep reinforcement learning.
\newblock In {\em ICML}, 2016.

\bibitem{newell2017pixels}
Alejandro Newell and Jia Deng.
\newblock Pixels to graphs by associative embedding.
\newblock In {\em NeurIPS}, 2017.

\bibitem{niu2019recursive}
Yulei Niu, Hanwang Zhang, Manli Zhang, Jianhong Zhang, Zhiwu Lu, and Ji-Rong
  Wen.
\newblock Recursive visual attention in visual dialog.
\newblock In {\em CVPR}, 2019.

\bibitem{norcliffe2018learning}
Will Norcliffe-Brown, Stathis Vafeias, and Sarah Parisot.
\newblock Learning conditioned graph structures for interpretable visual
  question answering.
\newblock In {\em NeurIPS}, 2018.

\bibitem{omidshafiei2017deep}
Shayegan Omidshafiei, Jason Pazis, Christopher Amato, Jonathan~P How, and John
  Vian.
\newblock Deep decentralized multi-task multi-agent reinforcement learning
  under partial observability.
\newblock In {\em ICML}, 2017.

\bibitem{qi2019attentive}
Mengshi Qi, Weijian Li, Zhengyuan Yang, Yunhong Wang, and Jiebo Luo.
\newblock Attentive relational networks for mapping images to scene graphs.
\newblock In {\em CVPR}, 2019.

\bibitem{qian2019video}
Xufeng Qian, Yueting Zhuang, Yimeng Li, Shaoning Xiao, Shiliang Pu, and Jun
  Xiao.
\newblock Video relation detection with spatio-temporal graph.
\newblock In {\em ACM MM}, 2019.

\bibitem{ranzato2016sequence}
Marc'Aurelio Ranzato, Sumit Chopra, Michael Auli, and Wojciech Zaremba.
\newblock Sequence level training with recurrent neural networks.
\newblock In {\em ICLR}, 2016.

\bibitem{rao2018learning}
Yongming Rao, Dahua Lin, Jiwen Lu, and Jie Zhou.
\newblock Learning globally optimized object detector via policy gradient.
\newblock In {\em CVPR}, 2018.

\bibitem{redmon2017yolo9000}
Joseph Redmon and Ali Farhadi.
\newblock Yolo9000: better, faster, stronger.
\newblock In {\em CVPR}, 2017.

\bibitem{ren2015faster}
Shaoqing Ren, Kaiming He, Ross Girshick, and Jian Sun.
\newblock Faster r-cnn: Towards real-time object detection with region proposal
  networks.
\newblock In {\em NeurIPS}, 2015.

\bibitem{ren2017deep}
Zhou Ren, Xiaoyu Wang, Ning Zhang, Xutao Lv, and Li-Jia Li.
\newblock Deep reinforcement learning-based image captioning with embedding
  reward.
\newblock In {\em CVPR}, 2017.

\bibitem{rennie2017self}
Steven~J Rennie, Etienne Marcheret, Youssef Mroueh, Jarret Ross, and Vaibhava
  Goel.
\newblock Self-critical sequence training for image captioning.
\newblock In {\em CVPR}, 2017.

\bibitem{shang2017video}
Xindi Shang, Tongwei Ren, Jingfan Guo, Hanwang Zhang, and Tat-Seng Chua.
\newblock Video visual relation detection.
\newblock In {\em ACM MM}, 2017.

\bibitem{shi2019explainable}
Jiaxin Shi, Hanwang Zhang, and Juanzi Li.
\newblock Explainable and explicit visual reasoning over scene graphs.
\newblock In {\em CVPR}, 2019.

\bibitem{simonyan2015very}
Karen Simonyan and Andrew Zisserman.
\newblock Very deep convolutional networks for large-scale image recognition.
\newblock In {\em ICLR}, 2015.

\bibitem{sutton1998reinforcement}
Richard~S Sutton, Andrew~G Barto, Francis Bach, et~al.
\newblock {\em Reinforcement learning: An introduction}.
\newblock MIT press, 1998.

\bibitem{sutton2000policy}
Richard~S Sutton, David~A McAllester, Satinder~P Singh, and Yishay Mansour.
\newblock Policy gradient methods for reinforcement learning with function
  approximation.
\newblock In {\em NeurIPS}, 2000.

\bibitem{tampuu2015multiagent}
Ardi Tampuu, Tambet Matiisen, Dorian Kodelja, Ilya Kuzovkin, Kristjan Korjus,
  Juhan Aru, Jaan Aru, and Raul Vicente.
\newblock Multiagent cooperation and competition with deep reinforcement
  learning.
\newblock In {\em arXiv}, 2015.

\bibitem{tang2019learning}
Kaihua Tang, Hanwang Zhang, Baoyuan Wu, Wenhan Luo, and Wei Liu.
\newblock Learning to compose dynamic tree structures for visual contexts.
\newblock In {\em CVPR}, 2019.

\bibitem{Wang_2019_CVPR}
Wenbin Wang, Ruiping Wang, Shiguang Shan, and Xilin Chen.
\newblock Exploring context and visual pattern of relationship for scene graph
  generation.
\newblock In {\em CVPR}, 2019.

\bibitem{weaver2001optimal}
Lex Weaver and Nigel Tao.
\newblock The optimal reward baseline for gradient-based reinforcement
  learning.
\newblock In {\em UAI}, 2001.

\bibitem{xu2017video}
Dejing Xu, Zhou Zhao, Jun Xiao, Fei Wu, Hanwang Zhang, Xiangnan He, and Yueting
  Zhuang.
\newblock Video question answering via gradually refined attention over
  appearance and motion.
\newblock In {\em ACM MM}, 2017.

\bibitem{xu2017scene}
Danfei Xu, Yuke Zhu, Christopher~B Choy, and Li Fei-Fei.
\newblock Scene graph generation by iterative message passing.
\newblock In {\em CVPR}, 2017.

\bibitem{xu2015show}
Kelvin Xu, Jimmy Ba, Ryan Kiros, Kyunghyun Cho, Aaron Courville, Ruslan
  Salakhudinov, Rich Zemel, and Yoshua Bengio.
\newblock Show, attend and tell: Neural image caption generation with visual
  attention.
\newblock In {\em ICML}, 2015.

\bibitem{xu2018dual}
Ning Xu, An-An Liu, Yongkang Wong, Yongdong Zhang, Weizhi Nie, Yuting Su, and
  Mohan Kankanhalli.
\newblock Dual-stream recurrent neural network for video captioning.
\newblock {\em CSVT}, 2018.

\bibitem{yang2018graph}
Jianwei Yang, Jiasen Lu, Stefan Lee, Dhruv Batra, and Devi Parikh.
\newblock Graph r-cnn for scene graph generation.
\newblock In {\em ECCV}, 2018.

\bibitem{yang2019auto}
Xu Yang, Kaihua Tang, Hanwang Zhang, and Jianfei Cai.
\newblock Auto-encoding scene graphs for image captioning.
\newblock In {\em CVPR}, 2019.

\bibitem{yang2018shuffle}
Xu Yang, Hanwang Zhang, and Jianfei Cai.
\newblock Shuffle-then-assemble: learning object-agnostic visual relationship
  features.
\newblock In {\em ECCV}, 2018.

\bibitem{yang2019learning}
Xu Yang, Hanwang Zhang, and Jianfei Cai.
\newblock Learning to collocate neural modules for image captioning.
\newblock In {\em ICCV}, 2019.

\bibitem{yao2018exploring}
Ting Yao, Yingwei Pan, Yehao Li, and Tao Mei.
\newblock Exploring visual relationship for image captioning.
\newblock In {\em ECCV}, 2018.

\bibitem{ye2017video}
Yunan Ye, Zhou Zhao, Yimeng Li, Long Chen, Jun Xiao, and Yueting Zhuang.
\newblock Video question answering via attribute-augmented attention network
  learning.
\newblock In {\em SIGIR}, 2017.

\bibitem{yin2018zoom}
Guojun Yin, Lu Sheng, Bin Liu, Nenghai Yu, Xiaogang Wang, Jing Shao, and
  Chen~Change Loy.
\newblock Zoom-net: Mining deep feature interactions for visual relationship
  recognition.
\newblock In {\em ECCV}, 2018.

\bibitem{yu2017joint}
Licheng Yu, Hao Tan, Mohit Bansal, and Tamara~L Berg.
\newblock A joint speakerlistener-reinforcer model for referring expressions.
\newblock In {\em CVPR}, 2017.

\bibitem{zellers2018neural}
Rowan Zellers, Mark Yatskar, Sam Thomson, and Yejin Choi.
\newblock Neural motifs: Scene graph parsing with global context.
\newblock In {\em CVPR}, 2018.

\bibitem{zhang2017visual}
Hanwang Zhang, Zawlin Kyaw, Shih-Fu Chang, and Tat-Seng Chua.
\newblock Visual translation embedding network for visual relation detection.
\newblock In {\em CVPR}, 2017.

\bibitem{zhang2017relationship}
Ji Zhang, Mohamed Elhoseiny, Scott Cohen, Walter Chang, and Ahmed Elgammal.
\newblock Relationship proposal networks.
\newblock In {\em CVPR}, 2017.

\bibitem{Zhang_2019_CVPR}
Ji Zhang, Kevin~J. Shih, Ahmed Elgammal, Andrew Tao, and Bryan Catanzaro.
\newblock Graphical contrastive losses for scene graph parsing.
\newblock In {\em CVPR}, 2019.

\bibitem{zhang2017actor}
Li Zhang, Flood Sung, Feng Liu, Tao Xiang, Shaogang Gong, Yongxin Yang, and
  Timothy~M Hospedales.
\newblock Actor-critic sequence training for image captioning.
\newblock In {\em NeurIPS Workshop}, 2017.

\bibitem{zhang2018learning}
Yan Zhang, Jonathon Hare, and Adam Pr{\"u}gel-Bennett.
\newblock Learning to count objects in natural images for visual question
  answering.
\newblock In {\em ICLR}, 2018.

\bibitem{zheng2015conditional}
Shuai Zheng, Sadeep Jayasumana, Bernardino Romera-Paredes, Vibhav Vineet,
  Zhizhong Su, Dalong Du, Chang Huang, and Philip~HS Torr.
\newblock Conditional random fields as recurrent neural networks.
\newblock In {\em ICCV}, 2015.

\bibitem{zhu2018deep}
Yaohui Zhu and Shuqiang Jiang.
\newblock Deep structured learning for visual relationship detection.
\newblock In {\em AAAI}, 2018.

\bibitem{zhuang2017towards}
Bohan Zhuang, Lingqiao Liu, Chunhua Shen, and Ian Reid.
\newblock Towards context-aware interaction recognition for visual relationship
  detection.
\newblock In {\em ICCV}, 2017.

\end{thebibliography}
}

\clearpage
\section*{Appendix}
\appendix

\setcounter{equation}{10}
\setcounter{figure}{6}

This supplementary document is organized as follows:
\begin{itemize}
    \item {Section \ref{sec:equ_details}} provides the details of some simplified functions in Agent Communication and Visual Relationship Detection.
    \item {Section \ref{sec:converge}} provides the detailed proof of the CMAT convergence, which guarantees that the proposed CMAT method can converge to a locally optimal policy.
    \item {Section \ref{sec:eq6}} provides the detailed derivation of Eq. (6), \ie, $\nabla_{\theta} J \approx \sum^n_{i=1} \nabla_{\theta} \log \bm{p}^T_i(v^T_i|h^T_i;\theta)Q(H^T, V^T) $.
    \item {Section \ref{sec:qualitative}} shows more qualitative results of CMAT compared with the strong baseline MOTIFS~\cite{zellers2018neural} in SGDet setting.
\end{itemize}

\section{Details of Some Simplified Functions} \label{sec:equ_details}
We demonstrate the details of some omitted functions in Eq. (1), Eq. (2), Eq. (3) and Eq. (4). 
\subsection{$F_s$ and $F_e$ in \textbf{\texttt{Extract}} Module} 
\begin{equation}
\begin{aligned}
    \bm{h}^{t}_i & = \text{LSTM}(\bm{h}^{t-1}_i, [\bm{x}^{t}_i, \bm{e}^{t-1}_i]), \\
    \bm{s}^t_i & = \bm{s}^{t-1}_i + \bm{W}_h \bm{h}^{t}_i, \\
    v^t_i & \sim \bm{p}^{t}_i = \text{softmax}(\bm{s}^t_i), \\
    \bm{e}^{t}_i &= \textstyle{\sum_{\tilde{v}}} \bm{p}^{t}_i(\tilde{v}) \mathbf{E}[\tilde{v}],
\end{aligned}
\end{equation}
where $\bm{h}^t_i \in \mathbb{R}^h$ is the hidden state of LSTM, $\bm{x}^t_i \in \mathbb{R}^d$ is the time-step input feature and $\bm{s}^t_i \in \mathbb{R}^{|\mathcal{C}|}$ is the object class confidence. $\mathbf{E}[\tilde{v}] \in \mathbb{R}^e$ is the embedding of class label $\tilde{v} \in \mathcal{C}$ and $\bm{e}^{t}_i \in \mathbb{R}^e$ is the soft-weighted embedding of class label based on probabilities $\bm{p}^t_i$. $\bm{W}_h \in \mathbb{R}^{h \times |\mathcal{C}|}$ is a learnable matrix.  and $[,]$ is concatenate operation. 

\subsection{$F_{m*}$ in \textbf{\texttt{Message}} Module}
\begin{equation}
\begin{aligned}
\bm{m}^t_j = \bm{W}_u \bm{h}^t_j, \; \bm{m}^t_{ij} = \bm{W}_p\bm{h}^t_{ij}
\end{aligned}
\end{equation}
where $\bm{m}^t_j \in \mathbb{R}^h$ and $\bm{m}^t_{ij} \in \mathbb{R}^h$ is the unary message and pairwise message, respectively. $\bm{h}^t_{ij} \in \bm{R}^d$ is the pairwsie feature between agent $i$ and agent $j$. $\bm{W}_u \in \mathbb{R}^{h \times h}$, $\bm{W}_p \in \mathbb{R}^{d \times h}$ are learnable mapping matrices.

\subsection{$F_{att*}$ and $F_{u*}$ in \textbf{\texttt{Update}} Module}
\begin{equation}
\begin{aligned}
    u^t_j &= \bm{w}_u [\bm{h}^t_i, \bm{h}^t_j],\; \alpha^t_j = \exp(u^t_j) / \textstyle{\sum_k} \exp(u^t_k), \\
    u^t_{ij} &= \bm{w}_p [\bm{h}^t_i, \bm{h}^t_{ij}],\; \alpha^t_{ij} = \exp(u^t_{ij}) / \textstyle{\sum_k} \exp(u^t_{ik}) \\
    \bm{x}^{t+1}_i &= \bm{W}_x (\text{ReLU} (\bm{h}^t_i + \textstyle{\sum_j} \alpha^t_j \bm{m}^t_j +  \textstyle{\sum_j} \alpha^t_{ij} \bm{m}^t_{ij})) \\
    \bm{h}^{t+1}_{ij} &= \text{ReLU} (\bm{h}^t_{ij} + \bm{W}_s \bm{h}^{t+1}_i + \bm{W}_e \bm{h}^{t+1}_j)
\end{aligned}
\end{equation}
where $\alpha^t_j$ and $\alpha^t_{ij}$ are attention weights to fuse different messages, $\bm{w}_u \in \mathbb{R}^{2h}$, $\bm{w}_p \in \mathbb{R}^{h+d}$, $\bm{W}_x \in \mathbb{R}^{h\times d}$, $\bm{W}_s \in \mathbb{R}^{h\times d}$, and $\bm{W}_e \in \mathbb{R}^{h\times d}$ are learnable mapping matrices.

\subsection{$F_r$ in Visual Relationship Detection}
\begin{equation}
\begin{aligned}
    \bm{z}_i & = \bm{W}_o [\bm{h}^T_i, \mathbf{E}[v^T_i]], \; \bm{z}_j = \bm{W}_o [\bm{h}^T_j, \mathbf{E}[v^T_j]], \\
    \bm{p}_{ij} & = \text{softmax} ([\bm{z}_i,  \bm{z}_j] \odot \bm{W}_r \bm{z}_{ij} + \bm{w}_{v^T_i, v^T_j}), \\
    r_{ij} & = \arg \textstyle{\max_{r \in \mathcal{R}}} \bm{p}_{ij}(r),
\end{aligned}
\end{equation}
where $\bm{W}_o \in \mathbb{R}^{(h+e) \times z}$, $\bm{W}_r \in \mathbb{R}^{z\times 2z}$ are transformation matrices, $\bm{z}_{ij} \in \mathbb{R}^z$ is the predicated visual feature between agent $i$ and $j$, $\odot$ is a fusing function~\footnote{Different functions get comparable performance. In our experiments, we follow~\cite{zhang2018learning}: $\bm{x} \odot \bm{y} = \text{ReLU}(\bm{W}_x \bm{x} + \bm{W}_y \bm{y}) - (\bm{W}_x \bm{x} - \bm{W}_y \bm{y})^2$.}, and $\bm{w}_{v^T_i, v^T_j} \in \mathbb{R}^{|\mathcal{C}|}$ is the bias vector specific to head and tail labels as in~\cite{zellers2018neural}. 

\noindent\textbf{Predicate Visual Features $z_{ij}$.} For the predicate visual features, we used RoIAlign to pool the union box of subject and object, and resized the union box feature to $7\times7\times512$. Following~\cite{zellers2018neural, dai2017detecting}, we used a $14\times14\times2$ binary feature map to model the geometric spatial position of subject and object, with one channel per box. We applied two convolutional layers on this binary feature map and obtained a new $7\times7\times512$ spatial position feature map. We added this position feature map with the previous resized union box feature, and applied two fully-connected layers to obtain the final predicate visual feature.

\clearpage
\onecolumn

\section{Proof of the Convergence of CMAT}\label{sec:converge}
\begin{proof}
We denote $\pi_i$ as the policy of agent $i$, \ie,$\pi_i=\bm{p}^T_i$ and $\bm{\pi}$ as the joint policy of all agents, \ie, $\bm{\pi} = \{\bm{p}^T_1, ..., \bm{p}^T_n \}$. Then, the expected gradient of CMAT is given by (cf. Eq. (11)):
\begin{align}
\nabla_{\theta} J & = \mathbb{E}_{\bm{\pi}} \left[\sum^n_{i=1} \nabla_{\theta} \log \pi_i(v^T_i) A^i(H^T, V^T)  \right], \\
& = \mathbb{E}_{\bm{\pi}} \left[\sum^n_{i=1} \nabla_{\theta} \log \pi_i(v^T_i) (R(H^T, V^T) - b(H^T, V^T_{-i}))  \right]. \nonumber
\end{align}
where the expection $\mathbb{E}_{\bm{\pi}}$ is with respect to the state-action distribution induced by the joint policy $\bm{\pi}$, $b(H^T, V^T_{-i})$ is the counterfactual baseline in CMAT model, \ie, $ b(H^T, V^T_{-i}) = \sum \bm{p}^T_i(\tilde{v}^T_i)R(H^T, (V^T_{-i}, \tilde{v}^T_i)). $

First, consider the expected contribution of this counterfactual baseline $b(H^T, V^T_{-i})$,
\begin{equation}
\nabla_{\theta} J_b = \mathbb{E}_{\bm{\pi}} \left[\sum^n_{i=1} \nabla_{\theta} \log \pi_i(v^T_i) b(H^T, V^T_{-i})  \right].
\end{equation}

Let $d^{\bm{\pi}}(s)$ be the discounted ergodic state distribution as defined by ~\cite{sutton2000policy}:
\begin{align}
\nabla_{\theta} J_b = & \sum_s d^{\bm{\pi}}(s)\sum^n_{i=1} \sum_{V^T_{-i}} \bm{\pi}(V^T_{-i})
\sum_{v^T_i} \pi_i(v^T_i) \nabla_{\theta} \log \pi_i(v^T_i)b(H^T, V^T_{-i}) \\
= & \sum_s d^{\bm{\pi}}(s) \sum^n_{i=1} \sum_{V^T_{-i}} \bm{\pi}(V^T_{-i}) \sum_{v^T_i} \nabla_{\theta} \pi_i(v^T_i) b(H^T, V^T_{-i}) \nonumber \\
= & \sum_s d^{\bm{\pi}}(s) \sum^n_{i=1} \sum_{V^T_{-i}} \bm{\pi}(V^T_{-i}) b(H^T, V^T_{-i}) \nabla_{\theta} 1 \nonumber \\
= & 0 \nonumber
\end{align}

Thus, this counterfactual baseline does not change the expected gradient. The reminder of the expected policy gradient is given by:
\begin{align}
\nabla_{\theta} J & = \mathbb{E}_{\bm{\pi}} \left[\sum^n_{i=1} \nabla_{\theta} \log \pi_i(v^T_i) R(H^T, V^T)  \right], \\
& = \mathbb{E}_{\bm{\pi}} \left[\nabla_{\theta} \log \prod^n_{i=1} \pi_i(v^T_i) R(H^T, V^T)  \right].
\end{align}
Writing the joint policy into a product of the independent policies:
\begin{align}
    \bm{\pi}(V^T) = \prod^n_{i=1} \pi_i(v^T_i),
\end{align}
we have the standard single-agent policy gradient:
\begin{align}
    \nabla_{\theta}J = \mathbb{E}_{\bm{\pi}} \left[\nabla_{\theta} \log \bm{\pi}(V^T)R(H^T, V^T) \right].
\end{align}

Konda~\etal~\cite{konda2000actor} proved that this gradient converges to a local maximum of the expected return $J$, given that: 1) the policy $\bm{\pi}$ is differentiable, 2) the update timescales for $\bm{\pi}$ are sufficiently slow. Meanwhile, the parameterization of the policy (\ie, the single-agent joint-action learner is decomposed into independent policies) is immaterial to convergence, as long as it remains differentiable.
\end{proof}

\section{Derivation of Eq. (6)} \label{sec:eq6}
Based on the policy gradient theorem we provide the detailed derivation of Eq. (6) as follows. We denote the action sequence for agent $i$ as $\hat{A}_i = \{\hat{a}^1_i, \hat{a}^2_i, ..., \hat{a}^T_i \}$, and value function $V_{\theta}(\hat{A}_i)$ as the expected future reward of sequence $\hat{A}_i$. Then the gradient of agent $i$ is:
\begin{equation}
    \begin{aligned}
        \nabla_{\theta} J_i & = \frac{dV(\hat{A}_i)}{d\theta} = \frac{d}{d\theta} 
        \mathbb{E}_{\hat{A}_i \sim \pi^t_i(\hat{A}_i)} R(\hat{A}_i) \\
        & = \sum_{\hat{A}_i} \frac{d}{d\theta}\left[\pi^t_i(\hat{a}^1_i) \pi^t_i(\hat{a}^2_i|\hat{a}^1_i)..\pi^t_i(\hat{a}^T_i|\hat{a}^1_i..\hat{a}^{T-1}_i) \right]R(\hat{A}_i) \\
        & = \sum^T_{t=1} \sum_{\hat{A}_i} \pi^t_i(\hat{A}^{1..t-1}_i) \frac{d\pi^t_i(\hat{a}^t_i|\hat{A}^{1..t-1}_i)}{d\theta} \pi^t_i(\hat{A}^{t+1..T}_i|\hat{A}^{1..t}_i)R(\hat{A}_i) \\
        & = \sum^T_{t=1} \sum_{\hat{A}^{1..t}_i} \pi^t_i(\hat{A}^{1..t-1}_i) \frac{\pi^t_i(\hat{a}^t_i|\hat{A}^{1..t-1}_i)}{d\theta} 
        \sum_{\hat{A}^{t+1..T}_i} \pi^t_i(\hat{A}^{t+1..T}_i|\hat{A}^{1..t}_i) \sum^T_{\tau=1} r^{\tau}_i(\hat{a}^{\tau}_i; \hat{A}^{1..\tau-1}_i) \\
        & = \sum^T_{t=1} \sum_{\hat{A}^{1..t}_i} \pi^t_i(\hat{A}^{1..t-1}_i) \frac{\pi^t_i(\hat{a}^t_i|\hat{A}^{1..t-1}_i)}{d\theta} 
         \left[r^t(\hat{a}^t_i; \hat{A}^{1..t-1}_i) + \sum_{\hat{A}^{t+1..T}_i} \pi^t_i(\hat{Y}^{t+1..T}_i|\hat{Y}^{1..t}_i) \sum^T_{\tau=t+1} r^{\tau}_i (\hat{a}^{\tau}_i; \hat{A}^{1..\tau-1}_i) \right] \\
        & = \sum^T_{t=1} \mathbb{E}_{\hat{A}^{1..t-1}_i \sim \pi^t_i(\hat{A}^{1..t-1}_i)} \sum_{a^t \in \mathcal{A}} \frac{d \pi^t_i(a^t|\hat{A}^{1..t-1}_i)}{d\theta} Q(a^t; \hat{A}^{1..t-1}_i) \\
        & = \mathbb{E}_{\hat{A}_i\sim \pi^t_i(\hat{A}_i)} \sum^T_{t=1} \sum_{a^t \in \mathcal{A}} \frac{\pi^t_i(a^t|\hat{A}^{1..t-1}_i)}{d\theta} Q(a^t; \hat{A}^{1..t-1}_i)
    \end{aligned}
\end{equation}
Further, the gradient for agent $i$ can be simplified as: 
\begin{equation}
    \begin{aligned}
        \nabla_{\theta} J_i &= \mathbb{E} \left[\sum^T_{t=1}\sum_{a^t_i \in \mathcal{A}} \nabla_{\theta} \pi^t_i(a^t_i)Q(s^t_i, a^t_i) \right] \\
        & = \mathbb{E} \left[ \sum^T_{t=1}\sum_{a^t_i \in \mathcal{A}} \pi^t_i(a^t_i) \nabla_{\theta} \log \pi^t_i(a^t_i)Q(s^t_i, a^t_i) \right] \\ 
        & \approx \sum^T_{t=1} \nabla_{\theta} \log \pi^t_i(a^t_i)Q(s^t_i, a^t_i)
    \end{aligned}
\end{equation}
Therefore, for the time step $t$, the gradient for agent $i$ is $\nabla_{\theta} \log \pi^t_i(a^t_i)Q(s^t_i, a^t_i)$. For multi-agent in a cooperative environment, the state-action function $Q$ should estimate the reward based on the set of all agent state and actions, \ie, $Q(S^t, A^t)$. Then, the gradient for all agents is:
\begin{equation}
    \begin{aligned}
      \nabla_{\theta} J \approx \sum^n_{i=1} \nabla_{\theta} J_i = \sum^n_{i=1}  \nabla_{\theta} \log \pi^t_i(a^t_i|s^t_i)Q(S^t, A^t).
    \end{aligned}
\end{equation}

In our CMAT, we CMAT samples actions after $T$-round agent communication, and the action for agent $i$ is $\bm{v}^T_i$, the policy function is $\bm{p}^T_i$, and the state of agent is $\bm{h}^t$, \ie, $S^t = H^t, A^t = V^t$. Therefore, the gradient for the cooperative multi-agent in CMAT is:
\begin{equation}
\begin{aligned}
\nabla_{\theta} J \approx \sum^n_{i=1} \nabla_{\theta} \log \bm{p}^T_i(v^T_i|h^T_i;\theta)Q(H^T, V^T)
\end{aligned}
\end{equation}

\section{More Qualitative Results} \label{sec:qualitative}
\begin{figure*}[htbp]
	\centering
	\includegraphics[width=1\linewidth]{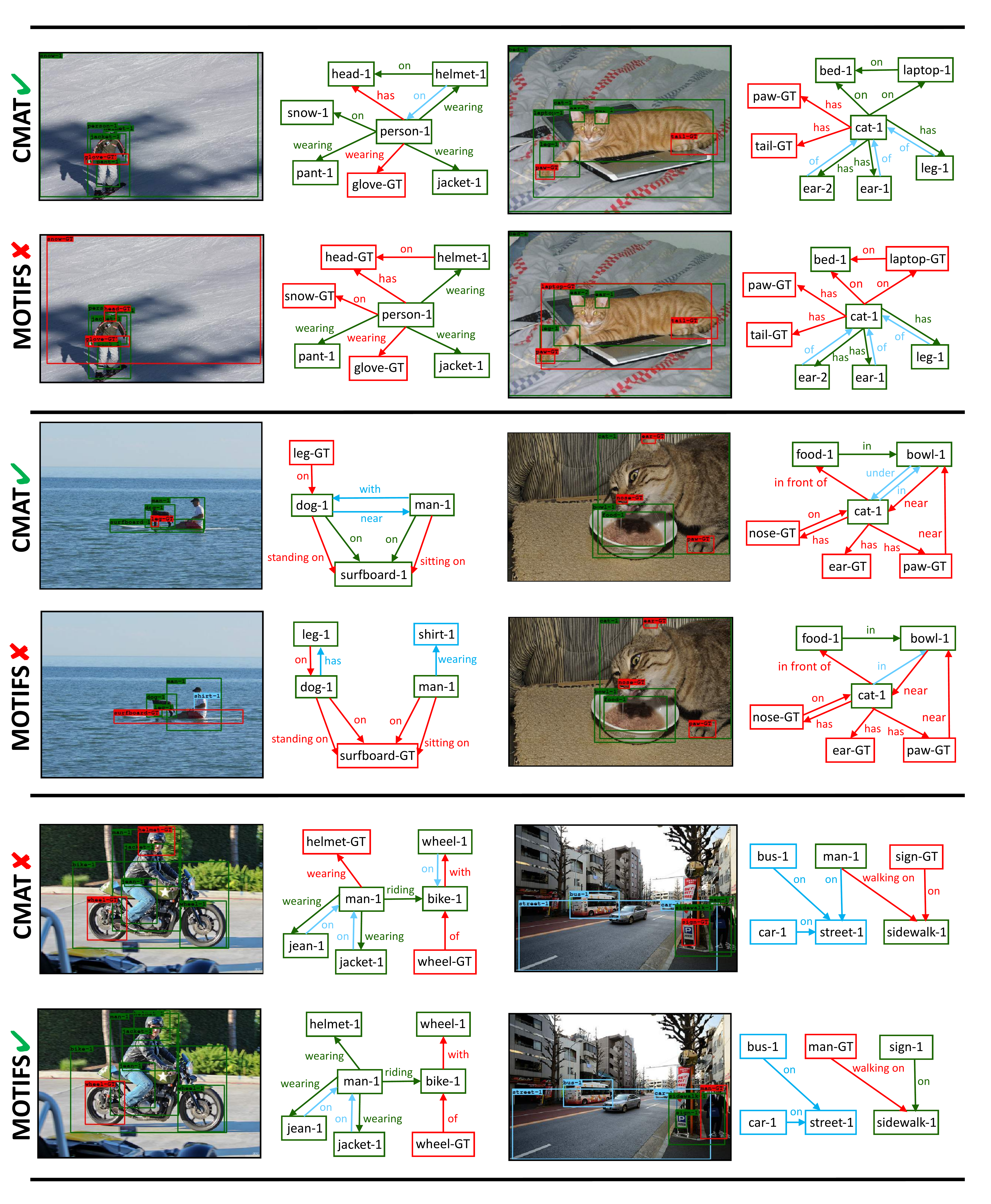}
	\caption{More qualitative results showing comparisons between CMAT and MOTIFS in the SGDet setting. Green boxes are detected boxes with IoU large 0.5 with the ground truth, blue boxes are detected but not labeled, red boxes are ground-truth with no match. Green edges are true positive predicted by each model at the R@20 setting, red edges are false negatives, and blue edges are false positives. Only detected boxes overlapped with ground-truth are shown.} 
\label{fig:7}
\end{figure*}

\begin{figure*}[htbp]
	\centering
	\includegraphics[width=1\linewidth]{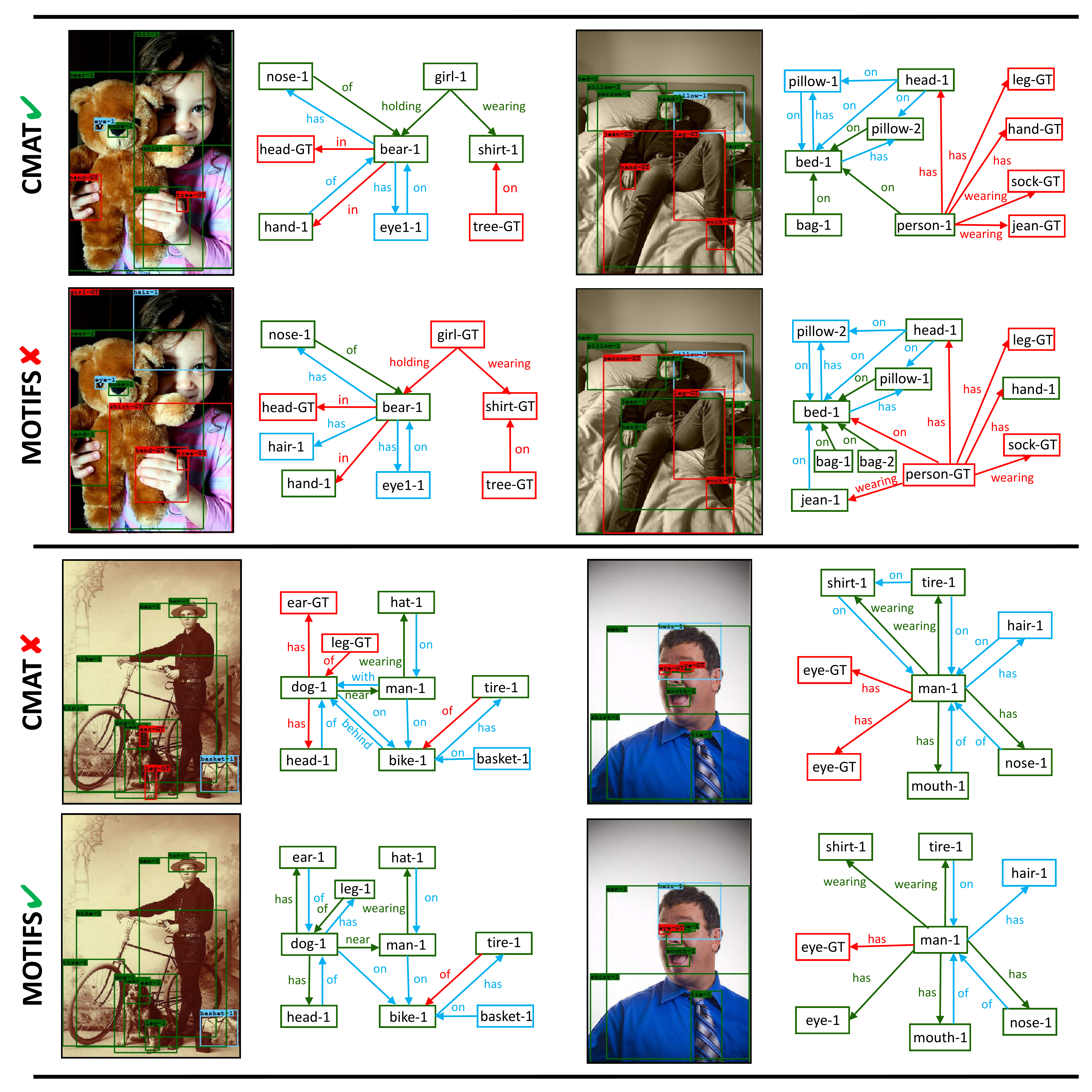}
	\caption{The caption is same as the one in Figure~\ref{fig:8}. More qualitative results showing comparisons between CMAT and MOTIFS in the SGDet setting. Green boxes are detected boxes with IoU large 0.5 with the ground truth, blue boxes are detected but not labeled, red boxes are ground-truth with no match. Green edges are true positive predicted by each model at the R@20 setting, red edges are false negatives, and blue edges are false positives. Only detected boxes overlapped with ground-truth are shown.}
\label{fig:8}
\end{figure*}

Figure~\ref{fig:7} and~\ref{fig:8} show more qualitative results of CMAT and MOTIFS in SGDet setting. From the rows where CMAT is better than MOTIFS, we can see that CMAT rarely mistakes at the important hub nodes such aas the ``surfboard" or ``laptop". This is because CMAT directly optimizes the graph-coherent objective. However, the rows show that the mistakes made by CMAT always come from the imcomplete anntation of CMAT can detect more false positive (the blue color) objects and relationship than MOTIFS. Since the evaluation metric (i.e., Recall@K) is based on the ranking of labeled triplet confidence, thus, detecting more reasonable false positive results with high confidence can worsen the performance.

\end{document}